\newcounter{myctr}

\documentclass{ws-ijhr}

\usepackage{url}
\usepackage{graphicx}
\usepackage{ifthen}
\usepackage{color}
\usepackage{gensymb}
\usepackage{soul}
\usepackage{enumitem}
\usepackage[framemethod=tikz]{mdframed}
\usepackage{refcount}

\graphicspath{{./figures/}}

\def\beg#1\eeg{\begin{gather}#1\end{gather}}
\renewcommand{\vec}[1]{\mathbf{#1}} 
\newcommand*{\fnref}[1]{\textsuperscript{\ref{#1}}}
\newcommand{\norm}[1]{\left\lVert#1\right\rVert}
\newcommand{\R}{\mathbb{R}}
\usepackage[font=scriptsize]{caption}
\usepackage{mwe}

\begin{document}

\markboth{D. Kanoulas et al.}{CoM-Based Grasp Pose Adaptation}

%
\catchline{}{}{}{}{}
%

\title{\large \bf Center-of-Mass-Based Grasp Pose Adaptation Using 3D
Range and Force/Torque Sensing}

\author{Dimitrios Kanoulas*, Jinoh Lee, Darwin G. Caldwell\\and Nikos G. Tsagarakis}

\address{Humanoids and Human-Centered Mechatronics Lab,\\ Istituto Italiano di Tecnologia (IIT)\\
Via Morego 30, 16163 Genova, Italy\\
*Dimitrios.Kanoulas@iit.it}

\maketitle

\begin{history}
\received{26 August 2016} %
\accepted{8 January 2018}  %
Published %
\end{history}

\begin{abstract}
Lifting objects, whose mass may produce high wrist torques that exceed the hardware strength limits, could lead to unstable grasps or serious robot damage.  This work introduces a new Center-of-Mass (CoM)-based grasp pose adaptation method, for picking up objects using a combination of exteroceptive 3D perception and proprioceptive force/torque sensor feedback.  The method works in two iterative stages to provide reliable and wrist torque efficient grasps.  Initially, a geometric object CoM is estimated from the input range data.  In the first stage, a set of hand-size handle grasps are localized on the object and the closest to its CoM is selected for grasping.  In the second stage, the object is lifted using a single arm, while the force and torque readings from the sensor on the wrist are monitored.  Based on these readings, a displacement to the new CoM estimation is calculated.  The object is released and the process is repeated until the wrist torque effort is minimized.  The advantage of our method is the blending of both exteroceptive (3D range) and proprioceptive (force/torque) sensing for finding the grasp location that minimizes the wrist effort, potentially improving the reliability of the grasping and the subsequent manipulation task.  We experimentally validate the proposed method by executing a number of tests on a set of objects that include handles, using the humanoid robot WALK-MAN.
\end{abstract}


\section{Introduction}
Grasping strategies for object manipulation have been extensively studied over the past few years,\cite{MS85,BicchiK00,LaValle06,KET07} especially for picking up light weight objects.  Recent advancement in control, planning, and perception have enabled robots to complete various manipulation tasks mostly considering the geometry of the object shape and largely excluding the weight and/or the mass distribution of the grasped/manipulated object.  Therefore, grasping, picking up, and eventually carrying heavy objects is considered an open problem and very challenging, particularly when the distribution of mass is not exteroceptively detectable.  Perception is a main aspect for completing these tasks, especially when grasp reliability during manipulation and efficient robot joints load reduction are required, while torque limitations also exist for the robotic arms.  In recent works, grasp detection has been usually achieved using either exteroceptive perception, such as 2D/3D visual\cite{tPP15} or tactile\cite{YBA11,JNMS15} sensing, where the mass distribution is not considered, or proprioceptive perception, such as force/torque sensing\cite{KVBP12} or the robot kinematics for contact detection.  Our approach uses a combination of exteroceptive and proprioceptive perception to improve grasping.

\begin{figure*}[b]
  \begin{center}
    \includegraphics[width=\textwidth]{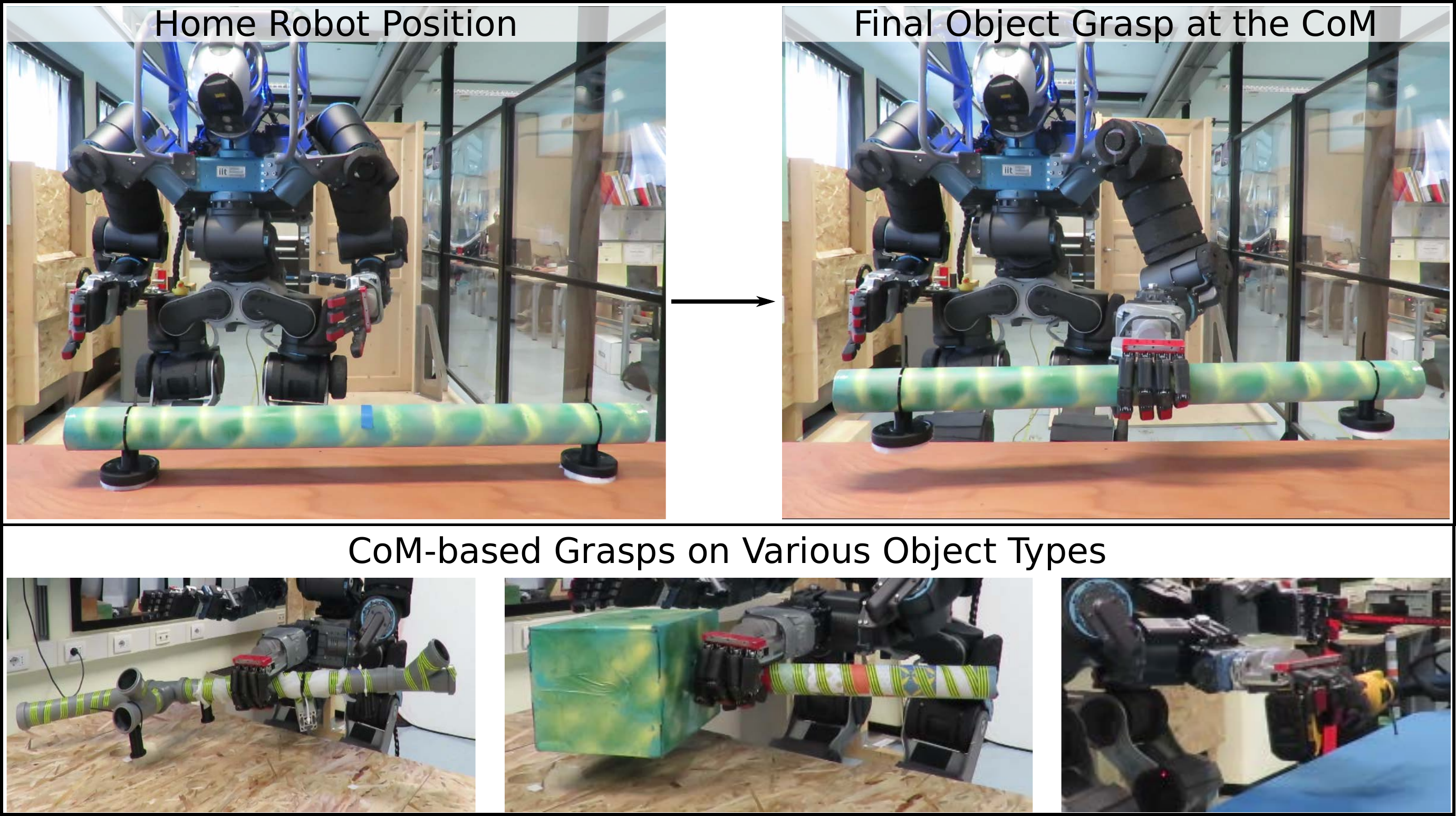}
  \end{center}
  \caption{The WALK-MAN humanoid robot and the proposed grasping for four different object types.}
  \label{Fig:robot}
\end{figure*}

Inspired from the debris task during the DARPA Robotics Challenge 2015, where wood debris pieces had to be removed from the robot's path, we present a new grasping method for a similar type of objects (Fig.~\ref{Fig:robot}).  The challenge when only exteroceptive perception is used is that the object weight and mass distribution are unknown, leading often to grasps that may generate high wrist moments.  These can eventually result in object drops (especially for underactuated hands) or robot overloading and instability.  The grasping force and wrist torque become the bottleneck in these scenarios, since there are strict hardware strength  force/torque limits.  To improve the reliability of such a grasp during manipulation, we present a novel method that combines 3D range and wrist force/torque sensing to detect the Center-of-Mass (CoM)-based grasp pose for objects which include handles and lie on support surfaces, like a tabletop.  In this paper, we study the case of single arm grasps, considering whole-body balancing.

\begin{figure*}[t]
  \begin{center}
    \includegraphics[width=\textwidth]{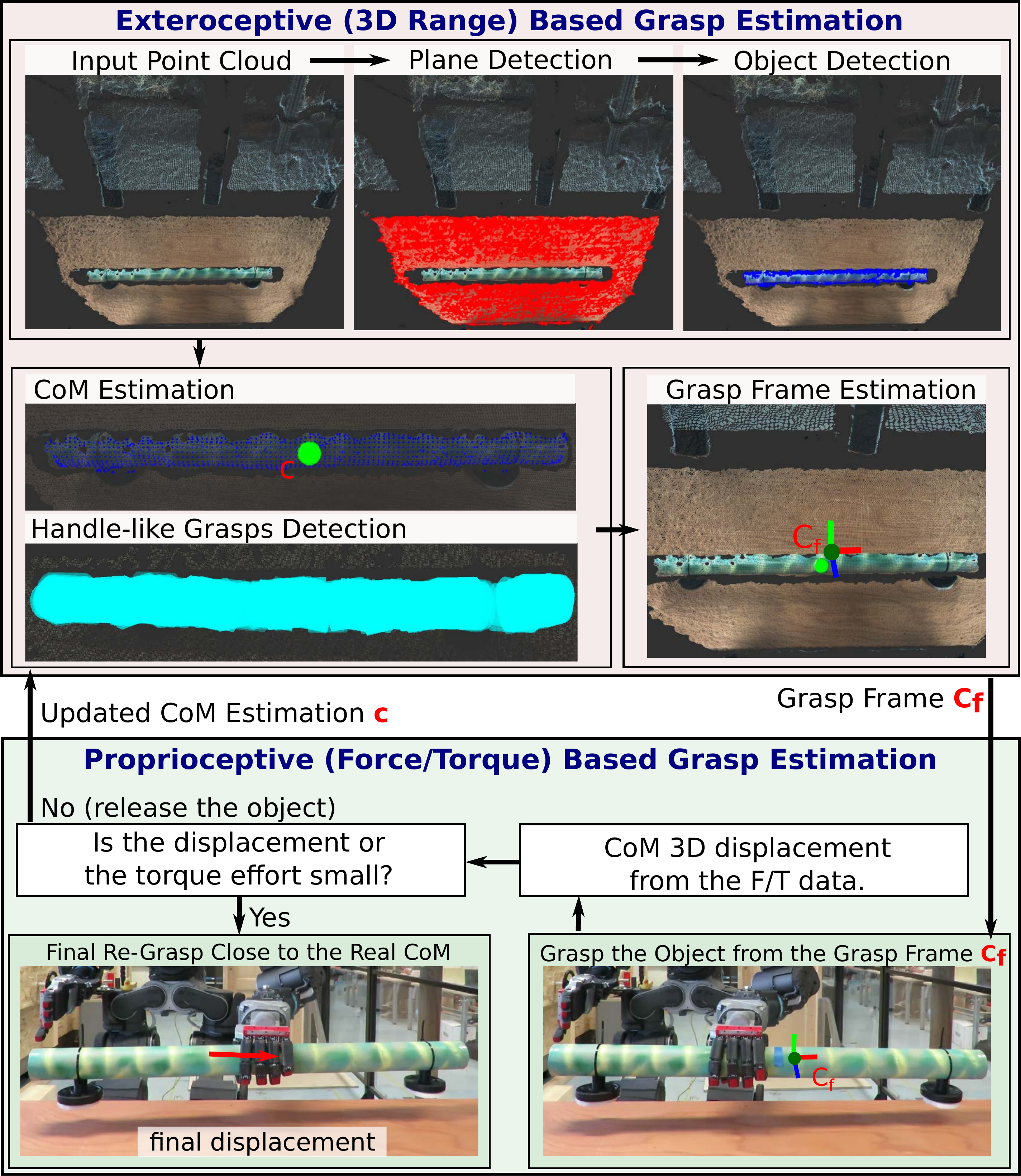}
  \end{center}
  \caption{System overview showing the main steps of the CoM-based grasp adaptation algorithm.}
  \label{Fig:algo}
\end{figure*}

The key aspect of our approach is the human-inspired hypothesis that holding heavy objects closer to their CoM makes the grasping more reliable and decreases the wrist torque effort when lifting them.  In addition, the risk of reaching the wrist torque limits is potentially reduced.  Our method is divided into two iterative stages (Fig.~\ref{Fig:algo}).  Initially, the object's CoM is estimated geometrically using voxelized 3D range data, under the assumption that it is made by isotropic material with constant mass density.  In the first exteroceptive-based stage, a set of handle-like (cylindrical) grasps are localized on the object.  From them, the closest to the estimated CoM is selected to grasp, minimizing the wrist torque effort among the several handle options (Sec.~\ref{Sec:perception}).  In the second proprioceptive-based stage, the robot lifts the object using a single arm, measuring in the meantime the wrist force/torque data.  A new CoM is estimated using these data, leading back to the first stage until a minimum wrist torque has been reached (Sec.~\ref{Sec:ft}).  It is worth noting that in our method, we achieve grasps, even when the actual CoM of the object does not lie in the object itself or inside a handle area.

Next, we review the research context, followed by a review of the handle-like grasp representation, as well as the robotic platform description (Sec.~\ref{Sec:algo}).  We then present in detail the 3D range exteroceptive perception system (Sec.~\ref{Sec:perception}) and the force/torque proprioceptive one (Sec.~\ref{Sec:ft}).  Finally, we present experimental results using the WALK-MAN humanoid robot\cite{Tsagarakis2016} on grasping heavy objects that include handles (Sec.~\ref{Sec:exp}).  The system is implemented in C++ using the Point Cloud\cite{RC11} and the Surface Patch\cite{SPL,Kanoulas14} Library, and is part of the source code designed for the DARPA Robotics Challenge Finals in 2015, used from the WALK-MAN team.

\subsection{Related work}\label{Sec:rw}
Object manipulation is considered one of the basic fields in robotics.\cite{MSZ94}  Both exteroceptive and proprioceptive perception were used to improve task completions.  Grasping involves either object picking/moving or tool manipulation tasks, where a tool needs to be grasped from a particular position to be used accordingly.

The area of picking up objects using perception was mainly focused on lightweight objects, considering mostly the geometry of the object shape and excluding the weight or mass distribution from the selection of the grasping point.  Recently, range sensing was extensively used for localizing grasps.  For instance, range data acquired from low cost structured light sensors were used to extract geometrically meaningful grasps such as cylindrical handles,\cite{tPP14} curved patches,\cite{KLTC17} or antipodal points\cite{tPP15} on light toy/kitchen objects to complete an empty-the-basket task.  Similarly, learning approaches on RGB-D data were used to improve grasping while clearing piles of toy objects,\cite{FV12,DEMK13,FVJ13} using also geometric representations such as rectangles.\cite{JMS11}  More recently, deep learning approaches were also considered in the literature, increasing the grasping success rate for similar type of objects,\cite{LLS15,Gualtieri16,PG16,LPKQ16} mainly applied on the Baxter robot which uses grippers.  Interactive approaches\cite{KKBS13,CSF12} were also used, where range data were checked for changes to verify and improve starting grasp hypotheses.  Geometric feature and template matching in 3D point clouds\cite{KRCGNK11,HPKRAS12,SSHB13} were also developed for grasp selection and planning.  All these methods were designed to work with high success rates for novel objects in clutter, using only range sensing.  The visual sensing limitations for heavier objects is been considered in our work, which defers from all the above ones.

Searching for tool object affordances in range and RGB images for a particular use, was also considered in the literature.\cite{PMPKRB08,VV12,PK13,Kaiser16,KKGMRMTA16}  Additionally, interaction between the robot and the environment also played a role in self-learning affordances.\cite{MLBS08,Mar2015}  More recently, deep learning was also used for pixelwise affordances classification.\cite{MTFA15,Nguyen16,Nguyen17}  Usually, tools such as drills and hammers need to be grasped in a particular way, which make the problem different than what we study in this paper.

Other types of sensing were also used to complete manipulation tasks.  For instance, tactile sensing\cite{PKTN06,LPYtPRSA14,OC01} was used for small object localization and manipulation (see Ref. \cite{YBA11} for a review).  Force/torque sensing was mainly used in the literature for the detection of contact,\cite{KVBP12} slippage,\cite{VBSKK13} or shape\cite{HCDDSD13} prediction.  An alternative approach to more stable grasping is through the object model learning.  For instance, tracking contacts while estimating the object's dimensions, mass, and friction\cite{ZLT13} or updating the object's attribute using tactile sensing\cite{LNK12} was studied in a probabilistic framework.  Towards active manipulation,\cite{Petrovskaya2016} object six Degrees-of-Freedom (DoF) localization takes place using various methods, such as the Scaling Series via touching\cite{PK11} or using Bayesian approaches for tactile sensing.\cite{VPBCN17}  In a much different direction, learning methods were also used for whole-body manipulation, for instance to learn friction models of the objects.\cite{SNK07}

Small object localization for in-hand manipulation was also extensively studied, using hybrid sensing methods.  In particular, vision, force, and tactile sensing was used in various sensing fusion combinations to compute both finger contacts and the applied forces during grasping.\cite{AMOL99}  Stereo vision and wrist force/torque sensing was used in combination with joint position sensing, to localize fingers with respect to object faces,\cite{HHMB11} while tactile and vision sensing was used to localize objects, providing robustness to occlusions and sensor failures for multi-fingered hands, in static configurations\cite{HHKM98} or during manipulation.\cite{BSAL13}

Task-oriented grasping methods using vision, proprioception, and
tactile sensing\cite{BSWK13} to increase stability were introduced in the direction of tool use.  Grasp adaptation in the controller level (stiffness) was also introduced\cite{LBKB14} to increase the performance, while regrasps were also used to reorient small objects in the environment.\cite{Nguyen16b}  Grasp planning was studied, such that grasps can guarantee lack of slippage and resistance to perturbations, using geometric object models and their theoretical CoM position and inertia.\cite{LDSA05}  CoM estimation from wrist force/torque data without the use of vision has an early history in the literature.\cite{AAH85}  In the closest work to our method,\cite{BYDK11,KP14} learning techniques for tactile coupled with vision sensing were developed to make a grasp more stable.  Still regrasping was not studied in any of these works.  To our knowledge, we are the first to use 3D range and wrist force/torque sensing iteratively to regrasp heavy objects with irregular mass distribution, based on the estimated CoM position for more stable and torque efficient manipulation.

\section{Grasp Representation and the Robotic System}\label{Sec:algo}
The goal of our method is the detection of a reliable and torque efficient grasp.  Such a grasp pose should be as close as possible to the object's CoM, where the applied torque is the minimum. The final grasp localization is estimated in two iterative stages until a termination threshold criterion is met, using exteroceptive and proprioceptive perception.  Note that for simplicity and clarity, in the next two sections, we present some of the results on a simple cylindrical object, but later in the experimental section, we show that the method is generic to any object that includes handles.  

Representing and localizing grasps in range data is well studied in robotic manipulation.  We are interested in the problem of finding all the possible graspable areas on the object.  For this reason, we apply one of the state-of-the-art methods\cite{tPP14} that uses cylindrical handles to geometrically represent grasp affordances.  The original paper focuses on lifting light objects by grasping the closest handle on the object.  In this paper, we extend the idea by a more sophisticated CoM-based grasp selection.  First, we briefly review the handle-like grasp affordances representation and localization algorithm using 3D point clouds.  We then present the robotic platform including both the exteroceptive and proprioceptive sensing system and the robotic hand that has been used.

\subsection{Grasp representation and localization} \label{Sec:grasp}
A grasp handle (illustrated in cyan color in the figures) is represented as a cylindrical shell, which is a set of a fixed number of co-linear cylinders of different radii.  Each cylinder is parametrized by its centroid, major principal axis, and radius.

The localization algorithm works as follows: Initially a set of uniform 3D points are sampled from the cloud that was acquired from a range sensor.  For each of these points, a local spherical point cloud neighborhood of fixed size (between 2cm and 3cm) is extracted.  Then, a quadric surface is algebraically fitted to the neighborhood, using Taubin's normalization method.  Given the curvatures along the two principal axis coming from the fitting, only those neighborhoods below some parametrized thresholds are considered.  For each one of those, a cylinder is fitted, assuring non-collision with surfaces during the object manipulation (i.e., a gap without points around the cylinder).  Handles are fixed sets of co-linear cylinders that are checked against some manually parametrized thresholds of their centroids, principal axes, and radii distances that form the final enveloping grasp affordance set.  More details can be found in the original paper.\cite{tPP14}  In this work, as described in Sec.~\ref{Sec:perception}, we extract a set of handles on the object of interest that samples it uniformly random, but very densely.

\subsection{Robotic platform} \label{Sec:robot}
\begin{figure*}[t!]
  \begin{center}
    \includegraphics[width=\textwidth]{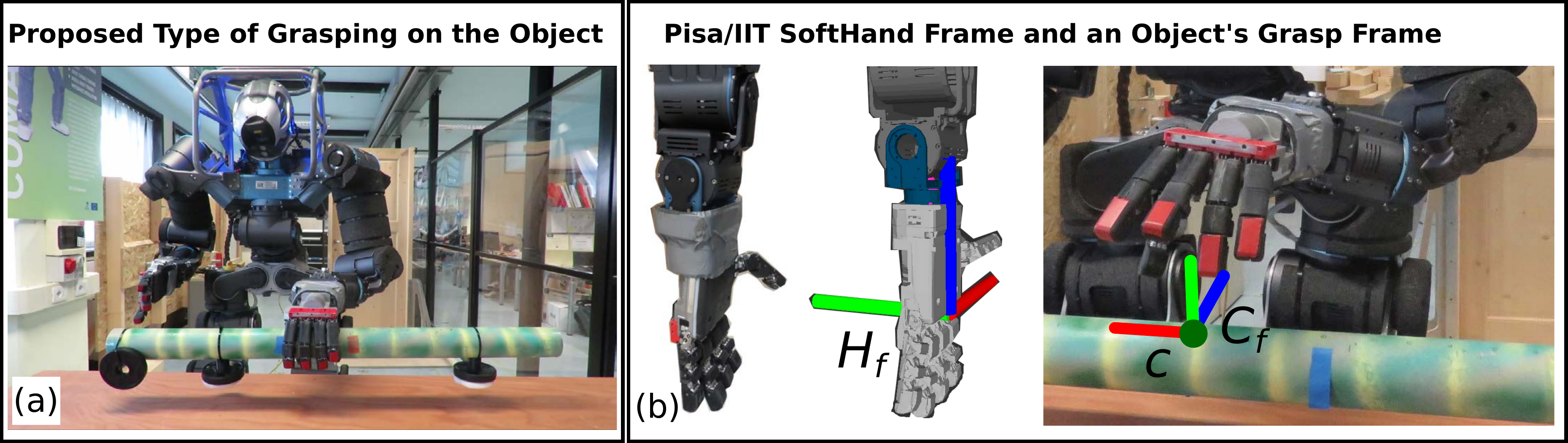}
  \end{center}
  \caption{(a) An object grasp with the WALK-MAN humanoid robot. (b) The real and the simulated Pisa/IIT SoftHand with its frame $H_f$ and a grasp frame $C_f$ with its origin point $\vec{c}$ on an object.}
  \label{Fig:grasp}
\end{figure*}
For the experiments, the WALK-MAN electrical motor driven humanoid robot has been used as shown in Fig.~\ref{Fig:grasp}(a).  WALK-MAN has 31 DoF, with two actuators for its hands, while it is 1.85m tall and weighs 118kg.  Visual sensing includes the CMU Multisense-SL system, which has a stereo and a LiDAR sensor, while four 6 DoF force/torque sensors are attached in the two wrists and ankles.  The hands are customized from the Pisa/IIT SoftHands\cite{CGSFPB12} of 11x11cm palm size and 12cm finger length.  During a grasp, the hand frame $H_f$ as appears in Fig.~\ref{Fig:grasp}(b), needs to co-align with the detected object's grasp frame $C_f$ at the origin point $\vec{c}$.  This grasp frame will be the handle frame that is closest to the CoM estimation.

There are three big challenges for the manipulation tasks using the particular robotic platform.  The first one has to do with the noisy stereo camera data, the second with the instabilities of the underactuated hand grasps, and the third with the robot's balancing during the execution of the task. These aspects are going to be considered in the following sections.  It is worth noting that grasp stability is importantly benefited from the active Pisa/IIT hand and passive WALK-MAN robot joint compliance.  Using the particular hardware helps with small grasp uncertainties either due to kinematic error accumulation or due to inaccurate grasp localization.  Moreover, we benefit from the finger compliance that enables the hand to envelop the contact surface, avoiding complicated control schemes.  When more dexterous manipulations are needed, for instance, manipulating smaller objects, the particular underactuated hand may be challenging to use.

\section{CoM-Based Grasp Pose Adaptation Method}
The introduced CoM-based grasp pose adaptation method includes two iterative stages (Fig.~\ref{Fig:algo}).  In the first one (Sec.~\ref{Sec:perception}), 3D visual handle grasps are localized on the object and the closest to the object's CoM is selected.  In the second stage (Sec.~\ref{Sec:ft}), the object is lifted and its CoM is estimated using the wrist force/torque data.  The stages are repeated until the minimum wrist torque effort is reached.

\subsection{Exteroceptive-based grasp estimation} \label{Sec:perception}
\begin{figure*}[t!]
  \begin{center}
    \includegraphics[width=\textwidth]{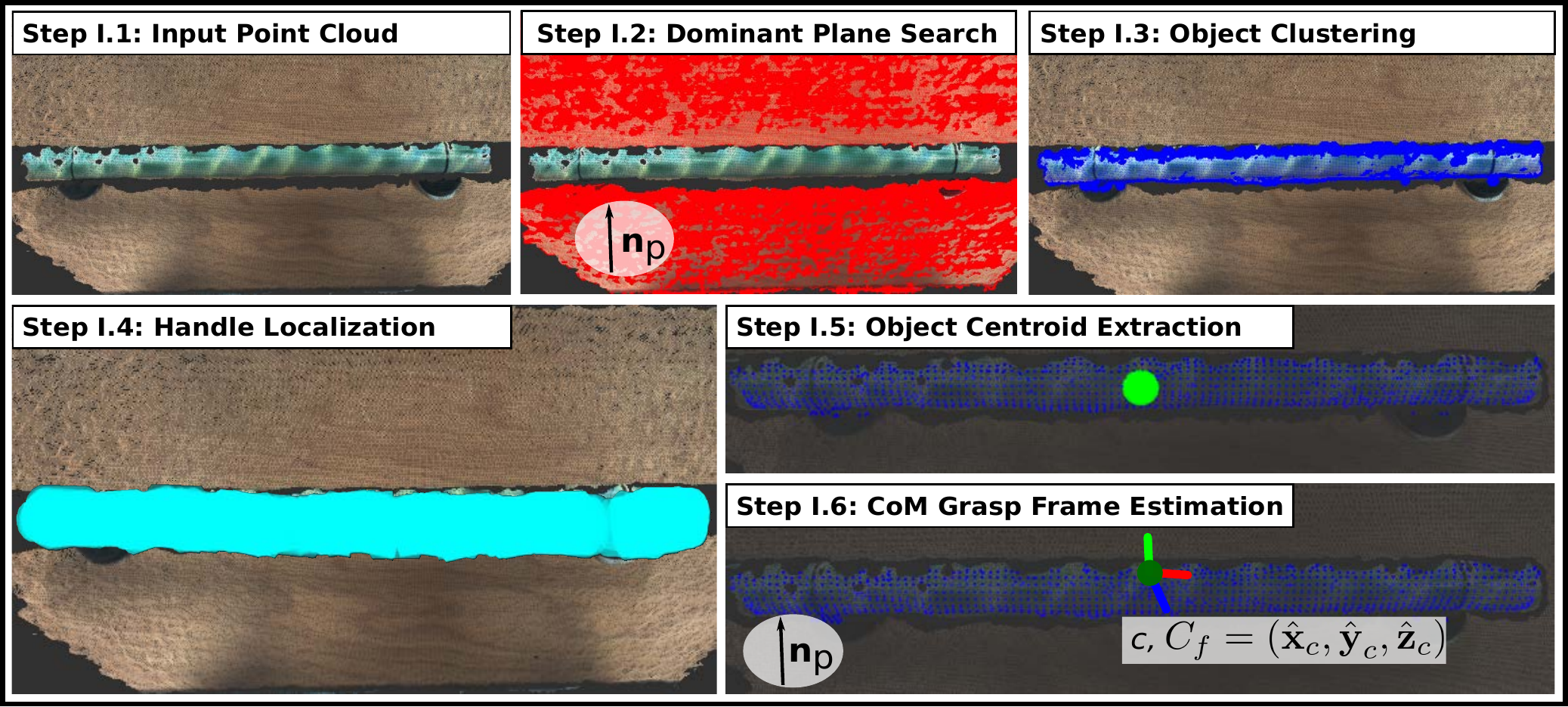}
  \end{center}
  \caption{The six steps of the 3D range grasp frame localization.  In red is the dominant plane, in cyan, the handle-like (cylindrical) grasps, in light green, the estimated CoM in the voxelized cloud, and in dark green the refined selected handle grasp position $\vec{c}$ with its frame $C_f$.}
  \label{Fig:perception-1}
\end{figure*}
To initially estimate the CoM, we use the 3D range point cloud data, acquired from the robot's range sensor.  In CAD systems, the CoM is detected by splitting the object into voxels and averaging the weighted distance from a fixed point, where the weights represent the object's density at the particular cube.\cite{KCY93}  Similarly, the proposed method \textit{visually} detects the initial CoM position of the object, using 3D voxelization and handle-grasps localization over the input object data, with the following steps, called `Stage I'. (see also Fig.~\ref{Fig:perception-1}):

\begin{mdframed}[userdefinedwidth=\textwidth]
\begin{description}
    \item[\underline{Stage I (3D range-based grasp localization)}]
\end{description}
\begin{itemize}[leftmargin=*]
  \item \textbf{Step I.1 [input point cloud]:} Acquire a point cloud from the range sensor.
  \item \textbf{Step I.2 [dominant plane search]:} Find the dominant plane (with its normal $\vec{n}_p$) and segment the points above it.
  \item \textbf{Step I.3 [object clustering]:} Cluster the segmented points into objects, using their Euclidean distances and then extract the largest one.
  \item \textbf{Step I.4 [handle localization]:} Fit a set of handle-like grasps of the size of the hand\cite{tPP14} to oversample the whole object cloud.
  \item \textbf{Step I.5 [object centroid extraction]:} Split the segmented object cloud space into fixed-size 3D voxels and for each one, compute its centroid; the median of the centroids represents the CoM position.
  \item \textbf{Step I.6 [CoM grasp frame estimation]:} From the extracted handle-like grasps, select the closest to the CoM estimation.
\end{itemize}
\end{mdframed}

\subsubsection{The algorithm}
The CMU Multisense-SL stereo camera provides an organized $1024\times1024$ point cloud in $15$Hz framerate.  After acquiring a cloud, we first filter out those points that are not reachable, to speed up the following computations.  These are the points that are approximately further than the distance that the hands can reach when they are in full extend, i.e., roughly $1.5$m away from the camera frame, given that the lower body is not used in this paper.

Assuming that the object lies on a flat surface, e.g., a tabletop, RANdom SAmpling Consensus (RANSAC) clustering procedure is used to extract the dominant plane cloud $P$ in the scene,\cite{HHRB11} with the angles between the point normals as the classification criterion.  The local normal vector for each point is computed using the integral images method\cite{HRDGN12} (i.e., covariance matrix estimation).  The extracted plane's normal vector towards the camera viewpoint will be denoted as $\vec{n_p}$.

To extract the points that are above the support surface in the direction of the normal vector $\vec{n_p}$, the plane cloud $P$ is first projected onto the fitted plane, a convex hull of the projected table points is created and all the rest points are then projected on the same plane.  For those that lie in the convex hull, we calculate the signed distance from the support surface (the positive is in the direction of $\vec{n_p}$), keeping finally only those with positive distance.  For these points, we apply a Euclidean clustering to extract the object clouds on the table.   In our scenario, we keep the largest cluster as the object to be grasped.  Note that in this stage, any object detection method can be applied, if a particular object needs to be grasped; also other types of clustering (e.g., using normals and curvatures) can be used in the place of the Euclidean to improve the segmentation.

For the first stage of the algorithm, to find the visually estimated CoM position, a 3D voxel grid of the object point cloud is created.  Each voxel is of fixed size and for each one, we replace the set of points that lies in it with their centroid.  The 3D voxelization is needed to be able to distribute equally the acquired points on the object.  Then, the CoM position is simply the median of the voxelized object point cloud.  Note that the estimated CoM may either be on a graspable area, a non-graspable one, or even outside the object.  For this reason, we also need to localize all the possible grasps on the object and select the closest to the CoM one.  This part of the method geometrically computes a CoM position, which is necessary for the initial object grasp.

A set of uniformly distributed cylindrical (quadratic curve) grasps are localized in real-time on the object of the size of the hand as described in Sec.~\ref{Sec:algo}.  The amount of grasps oversamples the object in a way such that there exists at least one grasp per centimetre in the graspable areas of the object.  A small cylindrical gap without points is guaranteed from the method, to accommodate for the grasping.  This part of the method makes it generic to all the objects that include handles.  From all these handle-like grasps, the closest to the estimated CoM is selected.  Its frame $C_f = (\hat{\vec{x}}_c, \hat{\vec{y}}_c, \hat{\vec{z}}_c)$ is defined as follows.  The $\hat{\vec{x}}_c$-axis is the cylinder axis pointing to the right, the $\hat{\vec{y}}_c$-axis is the unit normal vector $\vec{n_p}$, while the $\hat{\vec{z}}_c$-axis is uniquely defined as the cross-product between the $\hat{\vec{x}}_c$-axis and the $\hat{\vec{y}}_c$-axis towards the range sensor.  The origin point $\vec{c}$, which is initially defined as the center of the cylinder, is translated on the surface of the local point cloud neighborhood in the direction of the normal vector $\vec{n_p}$.  This CoM grasp frame $C_f$ at its origin $\vec{c}$ is the one that needs to co-align with the hand frame $H_f$ (see Fig.~\ref{Fig:grasp}(b)) during the grasping.

Point cloud filtering is an important step to make the fitting method\cite{tPP14} work on our stereo camera system.  The method was originally developed for the very accurate structured light Asus XTion sensor, which preserves the cylindrical geometry of a surface.  In contrary, our stereo camera point cloud is showing a big number of outliers and local spikes.  For this reason, a real-time statistical outlier removal and a second degree moving least squares filtering has been applied on the object point cloud.

\subsection{Proprioceptive-based grasp estimation} \label{Sec:ft}
\begin{figure*}[!b]
  \begin{center}
    \includegraphics[width=\textwidth]{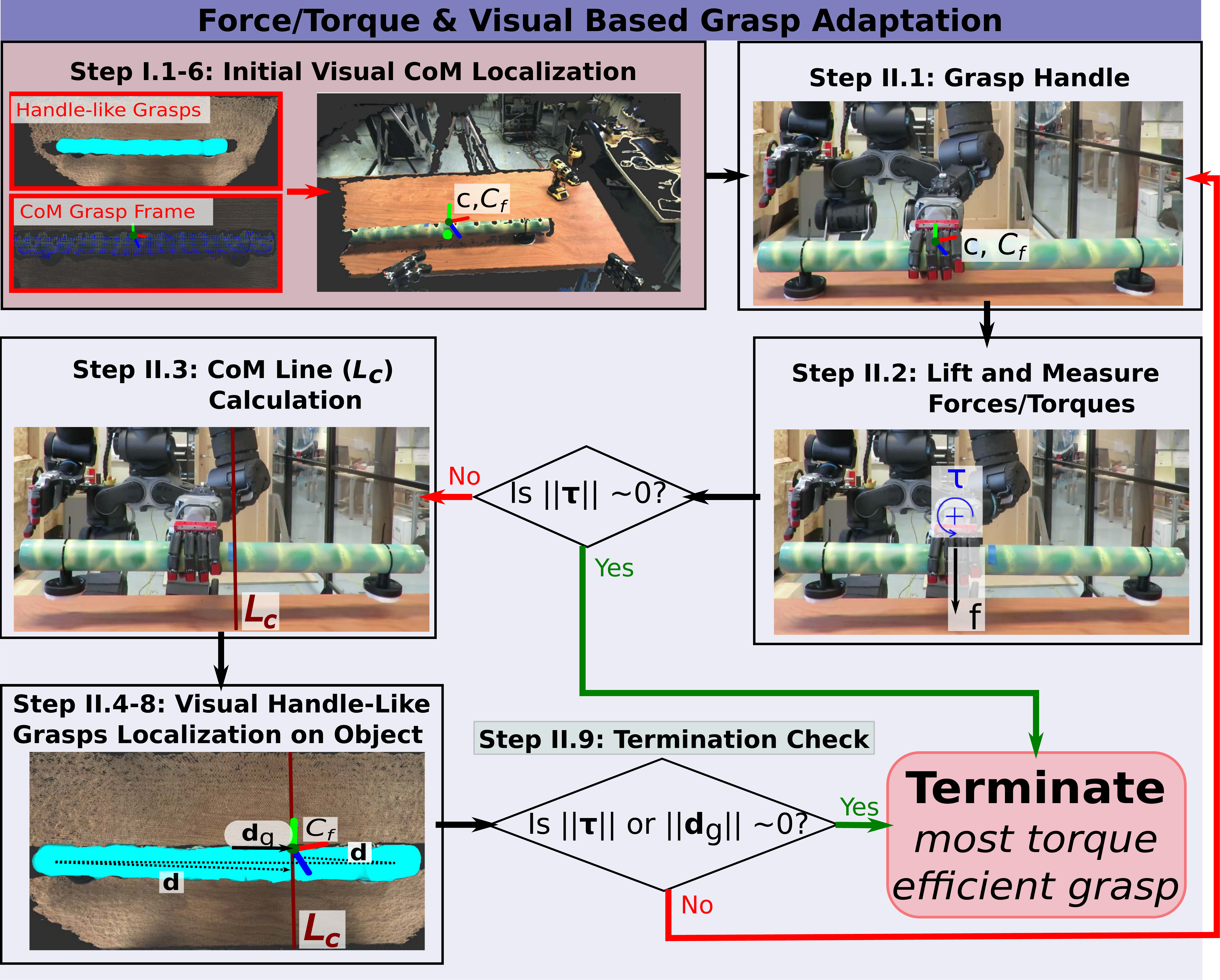}
  \end{center}
  \caption{The flowchart of the whole algorithm: the initial 3D range-based localization of the CoM (Stage I) and the wrist force/torque-based regrasping adaptation (Stage II).}
  \label{Fig:ft_com}
\end{figure*}

A visual CoM localization from an exteroceptive range sensor could provide a first grasp estimation.  Due to the point cloud data uncertainties (e.g., variations in points position, outliers, or missing areas due to occlusions) and a potential uneven distribution of mass along the object, the visually estimated CoM position may not be the same as the actual one.  Moreover, an exteroceptive range sensor is limited to represent only points on the surface of an object.  Using the F/T sensor, which is installed at the wrist level, a 6 DoF force and moment vector can be measured.  From these and the vision in the loop, the 3D displacement can be calculated, through a sequence of grasps and lifts, such that the wrist torque is minimized.  In particular, we follow the next steps, called `Stage II' (illustrated also as a flowchart, presented in Fig.~\ref{Fig:ft_com}):

\begin{mdframed}[userdefinedwidth=\textwidth]
\begin{description}
    \item [\underline{Stage II (force/torque-based grasp adaptation)}]
\end{description}
\begin{itemize}[leftmargin=*]
  \item \textbf{Step II.1 [grasp handle]:} Approach and grasp the object at the selected grasp handle.
  \item \textbf{Step II.2 [lift and measure forces/torques]:} Lift the object slightly and measure the forces and torques from the wrist F/T sensor.  If the minimum torque threshold has been reached, terminate.  Otherwise, lower and release the object.
  \item \textbf{Step II.3 [CoM line ($L_{c}$) calculation]:} Based on the forces and torques, calculate the \textit{CoM line} that goes through the CoM point of the object.
  \item \textbf{Step II.4--8 [visual handle-like grasps localization on object]} Run Steps I.1--4 (Sec.~\ref{Sec:perception}).  Select as the next handle grasp the one with the minimum torsional effort with respect to the CoM line.
  \item \textbf{Step II.9 [termination check]} If the minimum displacement distance or minimum torque has been reached, terminate after grasping and lifting the object.  Run Step I.6 (Sec.~\ref{Sec:perception}) and go to Step II.1.
\end{itemize}
\end{mdframed}

\subsubsection{The algorithm}
A final grasp may bring grasping instabilities for heavy objects with irregular mass distribution, when the actual CoM is far from the grasp point estimated only using the object geometry.  This is especially true when underactuated hands are used, like those of our robot, where object slips are unavoidable.  For this reason, the use of proprioceptive sensing may be essential to improve the initial estimation after the first visually driven grasp.

First, note that from the robot's kinematics, all the data can be transformed to a fixed world frame, and in the rest of the paper, it will be considered to be the Waist frame of the robot.  The initial input of this stage is the CoM point and the corresponding closest grasp frame $C_f = (\hat{\vec{x}}_c, \hat{\vec{y}}_c, \hat{\vec{z}}_c)$ at the origin position $\vec{c}$, that were estimated using the exteroceptive 3D visual perception method (Stage I in Sec.~\ref{Sec:perception}).  Initially, the hand approaches and grasps the visual contact point by co-aligning the hand frame $H_f$ with the grasp frame $C_f$ at $\vec{c}$ (Fig.~\ref{Fig:robot}(d)).  Then, the object is lifted slightly from the support surface.

While the object is lifted and it is not moving, the force $\vec{f} = (f_x, f_y, f_z)$ and the torque $\boldsymbol{\tau} = (\tau_x, \tau_y, \tau_z)$ vectors are measured at the wrist $F/T$ sensor (their values are averaged over time for two seconds). Based on the standard force/torque relation ($\boldsymbol{\tau} = \vec{d} \times  \vec{f}$), using the property of vector triple product, the distance vector $\vec{d}$ to the object's CoM is:
\beg \label{Eq:vecd}
  \vec{d} = \lambda \vec{f} + \frac{\vec{f} \times \boldsymbol{\tau}}{\norm{\vec{f}}^2}, \forall \lambda \in \R 
\eeg
where $\left\lVert \ \right\rVert$ is the vector norm.  This solution represents a set of vectors that go through the CoM point of the object and form a line $L_{c}$ parallel to the gravity vector.  If the torque effort, after the object is lifted, is smaller than a threshold, i.e., $\norm{\boldsymbol{\tau}} \leq \tau_{thres}$, it is assumed that the real CoM has been reached and no further action is required.

Otherwise, given the CoM line $L_{c}$, the object is lowered and released on the support surfaced and the visual stage is repeated as follows.  The object is segmented from the support surface and a new set of handle-like grasps are localized on it (Steps I.1-4). Then, these handle grasps are evaluated with respect to the line $L_{c}$ and their potential torque effort, such that the most wrist torque efficient one is selected.  To do so, the displacement vector $\vec{d}_h$ between each handle-like grasp frame (computed in Step I/II.6) and the CoM line $L_{c}$ needs to be calculated.  The problem of calculating the distance between a point and a line is a standard calculus problem.\cite{BJ52}  Assuming that the applied force $\vec{f}$ is only due to the object's mass towards the gravity, which was computed during the first object lift, the handle frame with the minimum potential torque effort, defined as the norm of the torque vector (i.e., $\norm{\boldsymbol{\tau}_h} = \norm{\vec{d}_h \times  \vec{f}}$), is selected.  Moreover, the corresponding displacement vector $\vec{d}_g$ to the new CoM estimation and its length $\norm{\vec{d}_g}$ are extracted.

If the new grasp frame displacement is smaller than a threshold distance $d_{thres}$, i.e., $\norm{\vec{d}_g} \leq d_{thres}$, it is assumed that the closest grasp to the real CoM point has been achieved and no further action is required.  Otherwise, the object is lowered and regrasped in the new displaced position, starting the second stage loop from the beginning.  The displacement threshold is required given that an object may not have a feasible grasp close to its CoM, and thus the torque effort threshold will not be efficient for the termination.

\section{EXPERIMENTS} \label{Sec:exp}
\begin{figure*}[!t]
  \begin{center}
    \includegraphics[width=\textwidth]{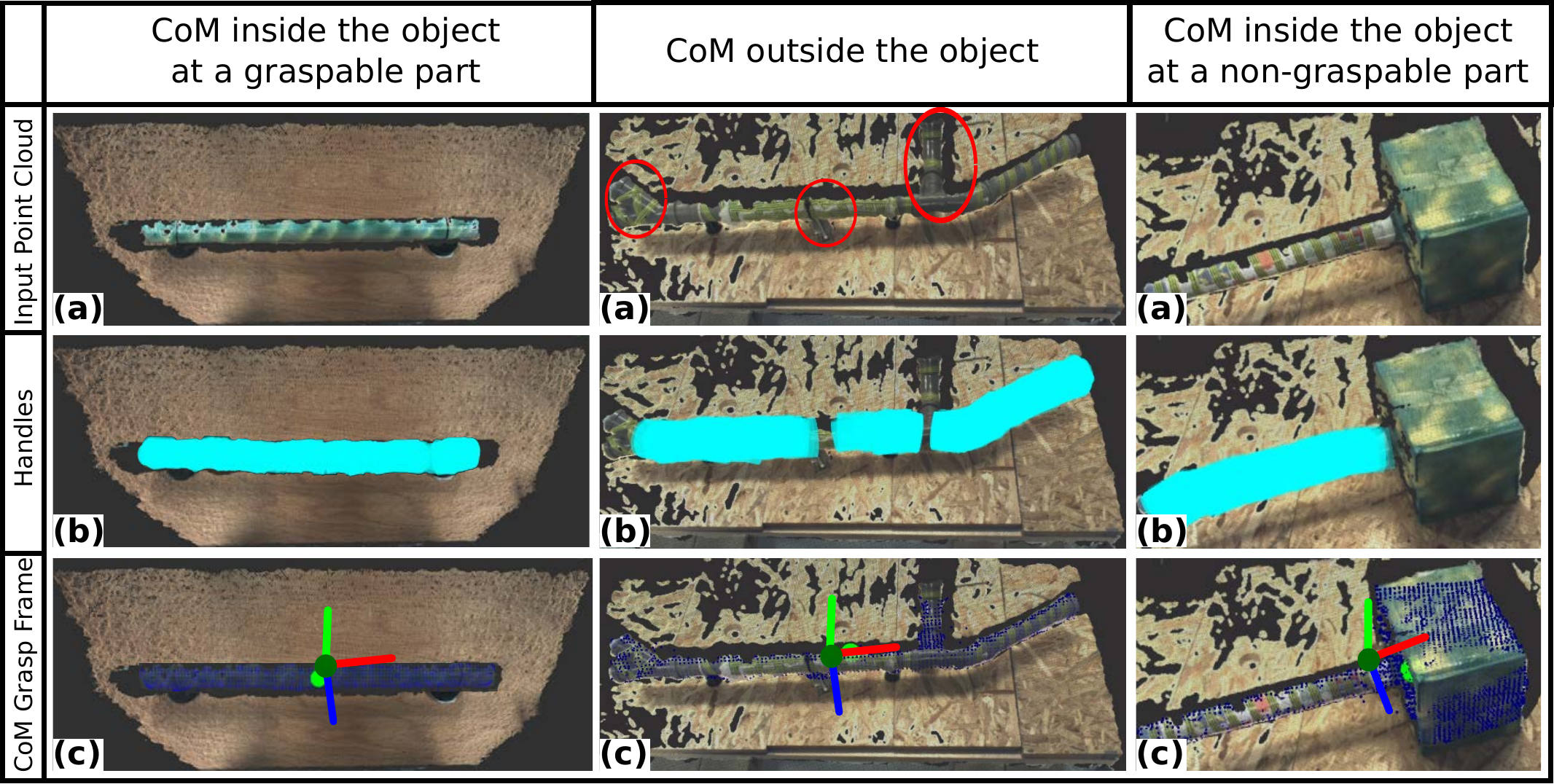}
  \end{center}
  \caption{The method for three object types: a handled one with its CoM at a graspable point, a partially handled one with its CoM outside the object, and one with its CoM at a non-graspable position.}
  \label{Fig:perception-2}
\end{figure*}

To test the overall approach, we run experiments on the humanoid robot WALK-MAN (introduced in Sec.~\ref{Sec:robot}).  We first test the ability to visually detect the CoM on various objects on a table.  Then, we test the regrasping process on three types of objects, by also changing their mass distribution: (i) a handle object that the real CoM is along its handle (Fig.~\ref{Fig:perception-2}---first column), (ii) a more complex object that the real CoM is outside the object and includes non-graspable areas (Fig.~\ref{Fig:perception-2}---second column), and finally (iii) an object that its CoM is inside the object, but in an non-graspable position (Fig.~\ref{Fig:perception-2}---third column).  We next discuss the hardware, control, and planning setup, as well as the experimental apparatus with the results.

\subsection{Control and Planning System}
The robot is controlled with the XBotCore\cite{MLHRCT17} and the YARP middleware framework,\cite{MFN06} while all the visual and force/torque perception data are handled with ROS.  Using the YARP functionalities, the high-level commands are created for the required motion primitives (e.g., ``reach'', ``grasp'', ``lift'', etc.) and delivered to the low-level torque controller, implemented on DSPs at each joints.\cite{SPVFHRMTB14}  In particular, to control the whole-body motion of the robot, inverse kinematics is resolved by the Stack-of-Task (SoT) formalism,\cite{EMW14} which employs cascaded Quadratic Programming (QP) solvers to efficiently find an optimum, in the least-square sense with a description of hierarchical tasks and constraints.  The OpenSoT control library has been used to provide these features.\cite{RHCT15}  Throughout the experiment, a single arm is controlled to handle the object manipulation depending on the grasp position.  The position of the other arm and the lower body are regulated, while the CoM of the full body is controlled to reside in the convex hull, for stable balancing during the task.

\subsection{Experimental Apparatus}
\begin{figure*}[!t]
  \begin{center}
    \includegraphics[width=0.7\textwidth]{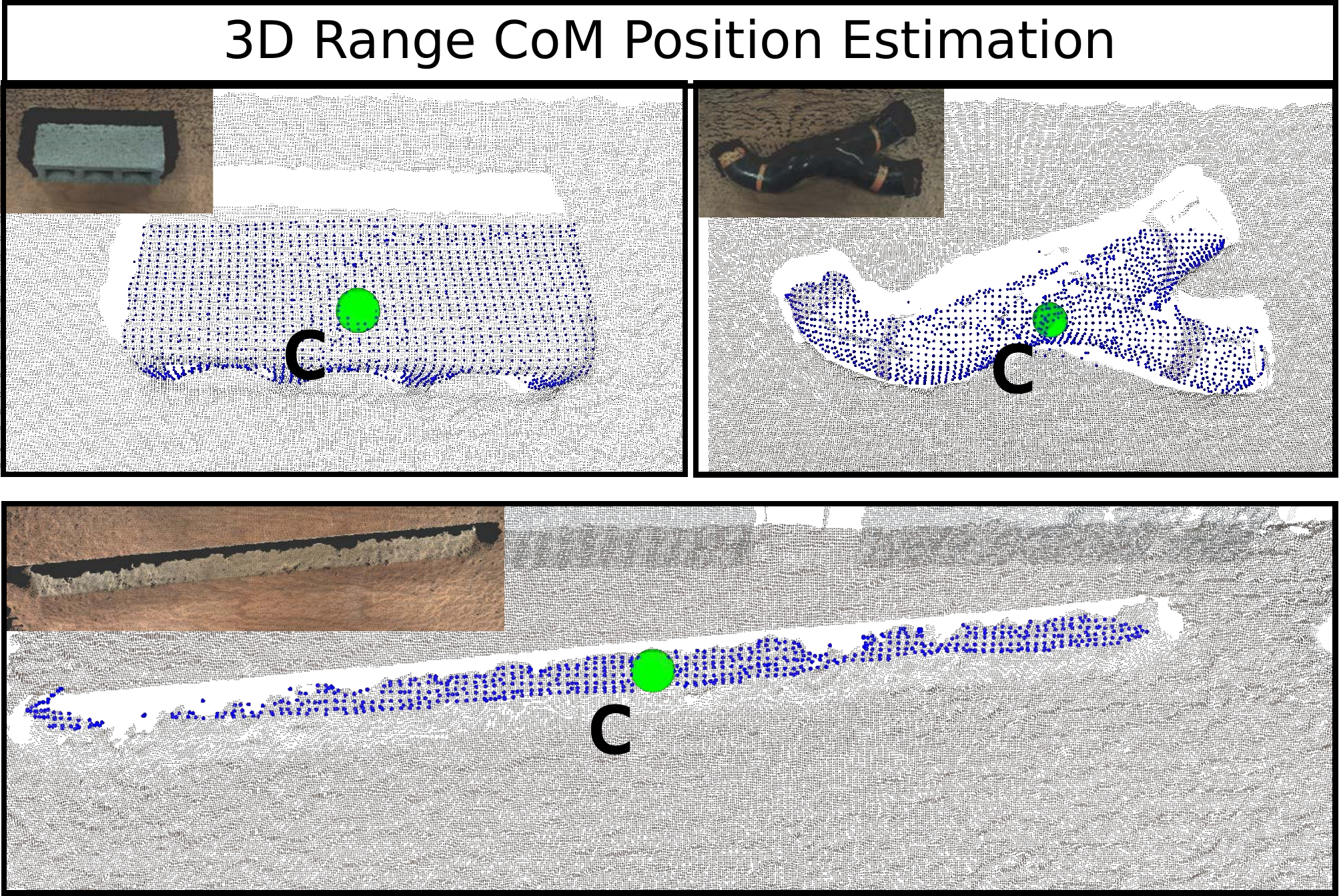}
  \end{center}
  \caption{A set of three objects and the estimated CoM (green dot) using 3D range voxelization.}
  \label{Fig:exp2}
\end{figure*}

\begin{figure*}[!t]
  \begin{center}
    \includegraphics[width=\textwidth]{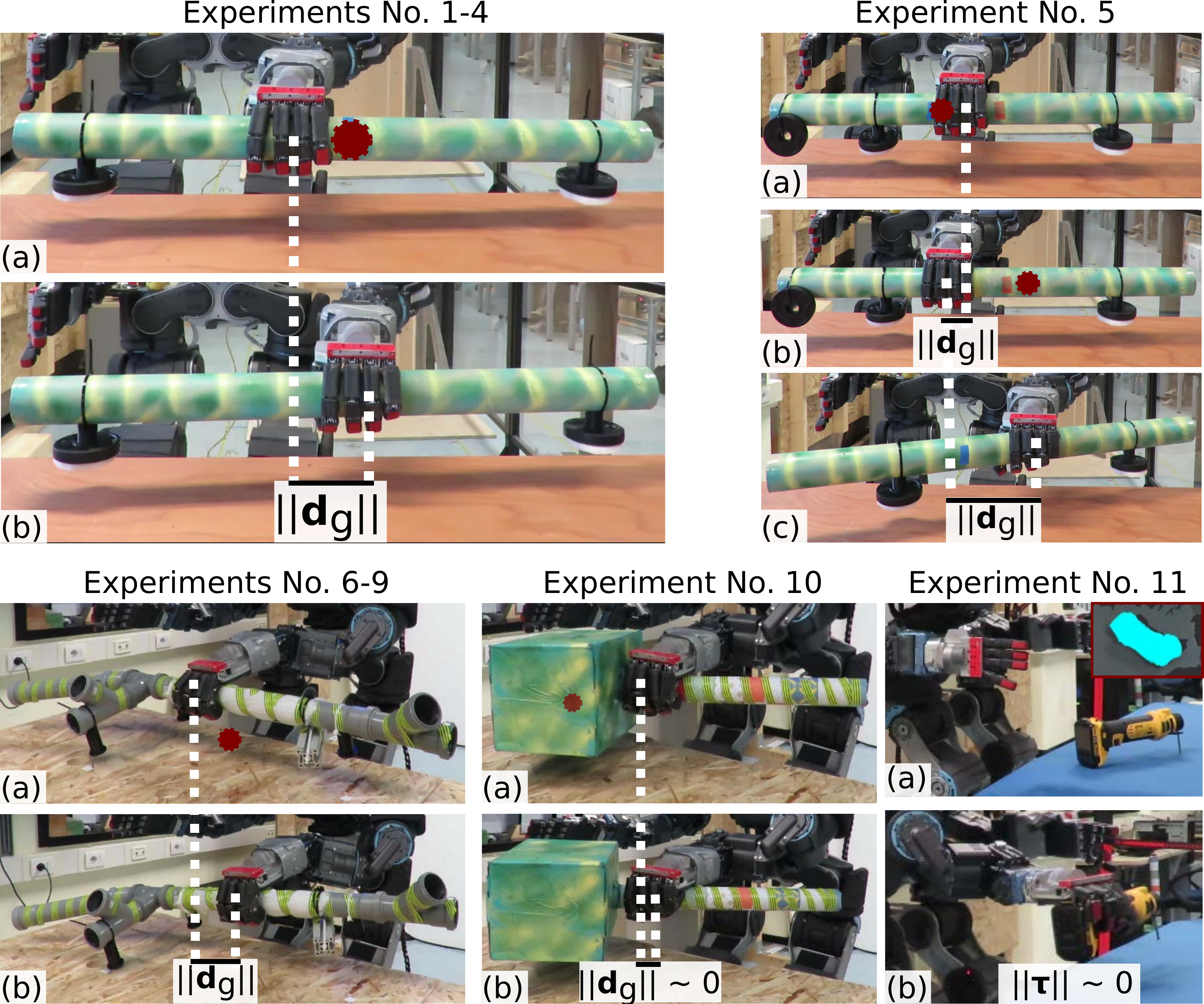}
  \end{center}
  \caption{Experimental instances of: (1) a handled object and the moving weights along its axis {(Exps. Nos. 1--4)}; (2) a handled object and the removed weight from its left part {(Exp. No. 5)}; (3) a partially-handled object, whose CoM is outside the object {(Exps. Nos. 6--9)}; (4) an object, whose CoM is at a non-graspable area {(Exp. No. 10)}.  Every sub-figure (a) visualizes the initial visual-based grasp and sub-figure (b) the minimum torque regrasp adaptation, while for Exp. No. 5, sub-figure (c) shows the regrasp adaptation after a weight is removed.  {Exp. No. 11} shows an experimental instance of a small handled object (drill), with the (a) localized handles and (b) it's minimum torque initial visual-based grasp.  The red star denotes the real CoM position.}
  \label{Fig:exp}
\end{figure*}

For the exteroceptive experimental testing, we ran the 3D range CoM estimation on various objects (handled or not) and evaluated qualitatively the results, some of which appear in Fig.~\ref{Fig:exp2}.

For the CoM-based grasp adaptation experiments, we set the robot in home position, $~50$cm in front of a flat $90$cm-tall table, where we place the objects in a reachable distance (see Fig.~\ref{Fig:robot}---upper left).  As shown in Fig.~\ref{Fig:exp}, the first object (Exps. Nos. 1--5\footnote{Only one experiment is visualized, while the rest can be found in the videos.\label{fn:exp}}; Fig.~\ref{Fig:perception-2}---first column) is a cylindrical debris of $7$cm diameter and $105$cm long, on which we attach $0.5$kg weights.  First, we attach two $0.5$kg weights, $10$cm distanced from each end (Exp. No. 1) and for each experiment we move the left weight $10$cm right (Exps. Nos. 1--4), changing in this way the position of the real CoM on the object (red dot).  We then add a third $0.5$kg weight (Exp. No. 5) and after the first regrasping we remove it manually to test the real-time online CoM position reestimation and the success of the regrasping according to the new measurements.  The second object (Exps. Nos. 6--9\fnref{fn:exp}; Fig.~\ref{Fig:perception-2}---second column) is a set of connected $5$cm diameter cylindrical parts.  Only some of them are graspable, while its actual CoM is not inside the object.  For each experiment we attach a $0.25$kg weight that each time we move it $10$cm left, along the white handle of the object.  The third object (Exp. No. 10; Fig.~\ref{Fig:perception-2}---third column) is a hammer-like one, where the CoM is inside its rectangle.  Last but not the least, we also tried our method on a small-handled object, i.e., a drill (Exp. No. 11).

\subsubsection*{Evaluation Measures}
For the first five experiments, we recorded the real CoM position $d_r$, the  visually calculated one $d_v$ (measured from the left most part of the object), the force $f_g$ along the gravity vector, the torque norm $\tau = \norm{\boldsymbol{\tau}}$ after the initial 3D visual grasp, the computed displacement distance $d = \norm{\vec{d}_g}$, the new torque $\tau'$, and the displacement value $d'$ after the new pose regrasp.  For the rest of the experiments (Exps. Nos. 6--10) we recorded the initial torque $\tau$ and the final one $\tau'$ after the regrasp.  Each experiment is performed $10$ times and Tables~\ref{tab:ExpResult1} and~\ref{tab:ExpResult2} present the average recorded measurements, while Figs.~\ref{Fig:expd} and~\ref{Fig:expft} visualize the CoM position deviations and torques for the initial proprioceptive and the reestimated extereoceptive graspings.  Note that in the beginning of each experiment, we remove all the force/torque sensor residuals, before grasping the object, while we manually set the thresholds $d_{thres} = 2$cm and $\tau_{thres}=~0.02$Nm.  All the experimental videos can be found under the following link:
\begin{center}
  \textbf{\url{http://dkanou.github.io/projects/com_grasping}}
\end{center}

\subsection{Results}
\begin{table*}[!h]
\begin{minipage}{\textwidth}
\caption{Average results of the first five experiments (Exps. Nos. 1--5).}
\label{tab:ExpResult1}
\begin{center}
\begin{tabular}{c|c|cccc|ccc}
\hline
\hline
    & real &\multicolumn{4}{c|}{3D visual CoM grasp}& \multicolumn{3}{c}{$F/T$ CoM regrasp}\\
No. & $d_r$(m) & $d_v$(m) & $f_g$(N) & $\tau$(Nm) & $d$(m) & $\tau^\prime$(Nm) & $d^\prime$(m) & $|\Delta \tau/\tau| (\%)$\\ 
\hline
1 & 0.520 & 0.450  & 12.0 & 0.886 & 0.0736 & -0.031  & -0.0020 & 96.5 \\
\hline
2 & 0.565 & 0.480  & 13.7 & 1.323 & 0.0969 & 0.090 & 0.0069 & 93.2 \\ 
\hline
3 & 0.600 & 0.490  & 15.9 & 1.734 & 0.1086 & 0.178 & 0.0177 & 89.7 \\ 
\hline
4 & 0.645 & 0.500  & 14.2 & 1.617 & 0.1138 & 0.147 & 0.0112 & 90.9 \\
\hline
5 & 0.360 & 0.390  & 17.9 & -0.648 & -0.0362 & 0.148 & 0.0084 & 22.8 \\ 
  & 0.530 & (0.390)& 14.9 & 1.514  & 0.1013  & 0.112 & 0.0082 & 92.6 \\
\hline
\hline
\end{tabular}
\end{center}
\end{minipage}
\end{table*}

\begin{table*}[!t]
\begin{minipage}{\textwidth}
\caption{Average results of the last five experiments (Exps. Nos. 6--10).}
\label{tab:ExpResult2}
\begin{center}
\begin{tabular}{c|c|cc}
\hline
\hline
    & Visual grasp & \multicolumn{2}{c}{F/T regrasp}\\
No. & $\tau$(Nm) & $\tau^\prime$(Nm) & $|\Delta \tau/\tau| (\%)$\\ 
\hline
6  & 0.460 & 0.1724 & 62.5\\
\hline
7  & 1.162 & 0.7075 & 39.1\\ 
\hline
8  & 1.096 & 0.7561 & 31.0\\ 
\hline
9  & 1.079 & 0.7153 & 33.7\\
\hline
10 & 1.330 & 0.538 & 59.5\\
\hline
\hline
\end{tabular}
\end{center}
\end{minipage}
\end{table*}

\begin{figure*}
  \begin{center}
    \includegraphics[width=0.7\textwidth]{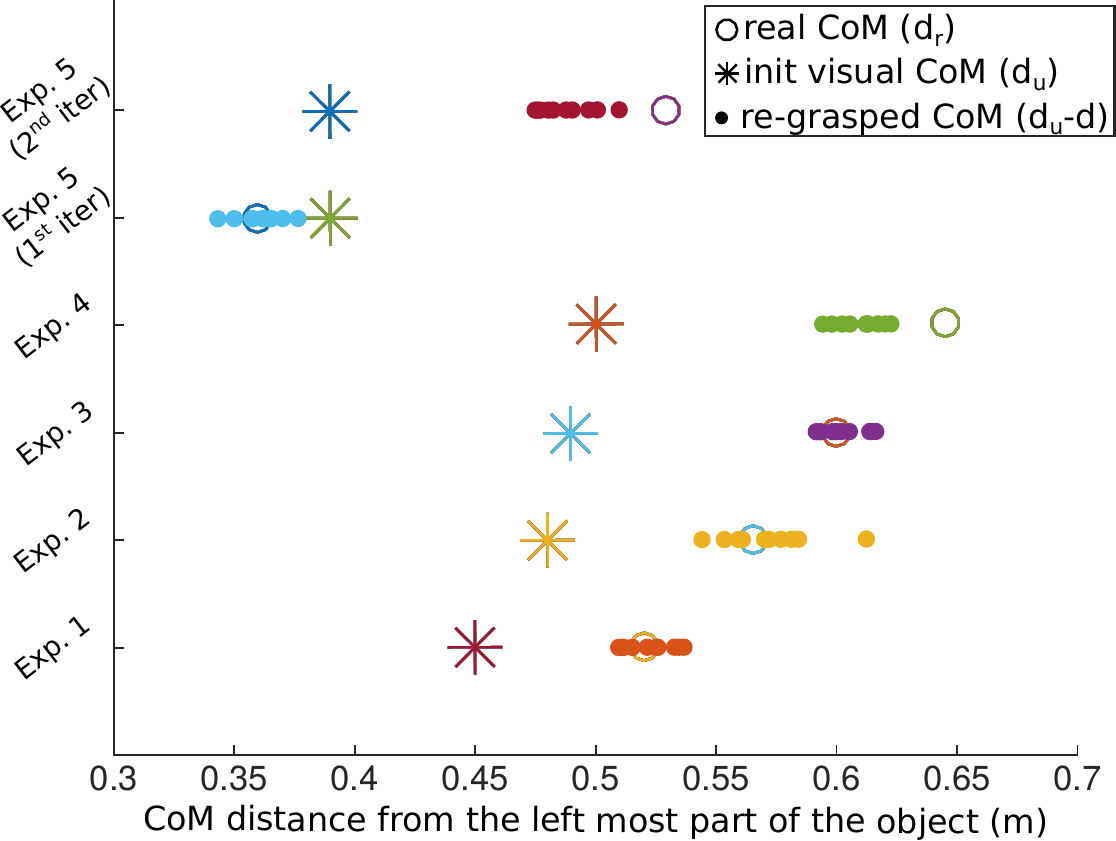}
  \end{center}
  \caption{For each of the $10$ repetitions of the Exps. Nos. 1--5, the initial visual CoM grasp position $d_u$, the regrasped adaptation displacement distance ($d_u-d$), and the real CoM position ($d_r$) are visualized.  All distances are measured from the left most part of the object.}
  \label{Fig:expd}
\end{figure*}

We first note that the visual system, with the handle-like grasp detection and the selection of the closest to the CoM estimation one, is working very reliably.  There was never noted any failure in the grasping.

\begin{figure*}
  \begin{center}
    \includegraphics[width=0.9\textwidth]{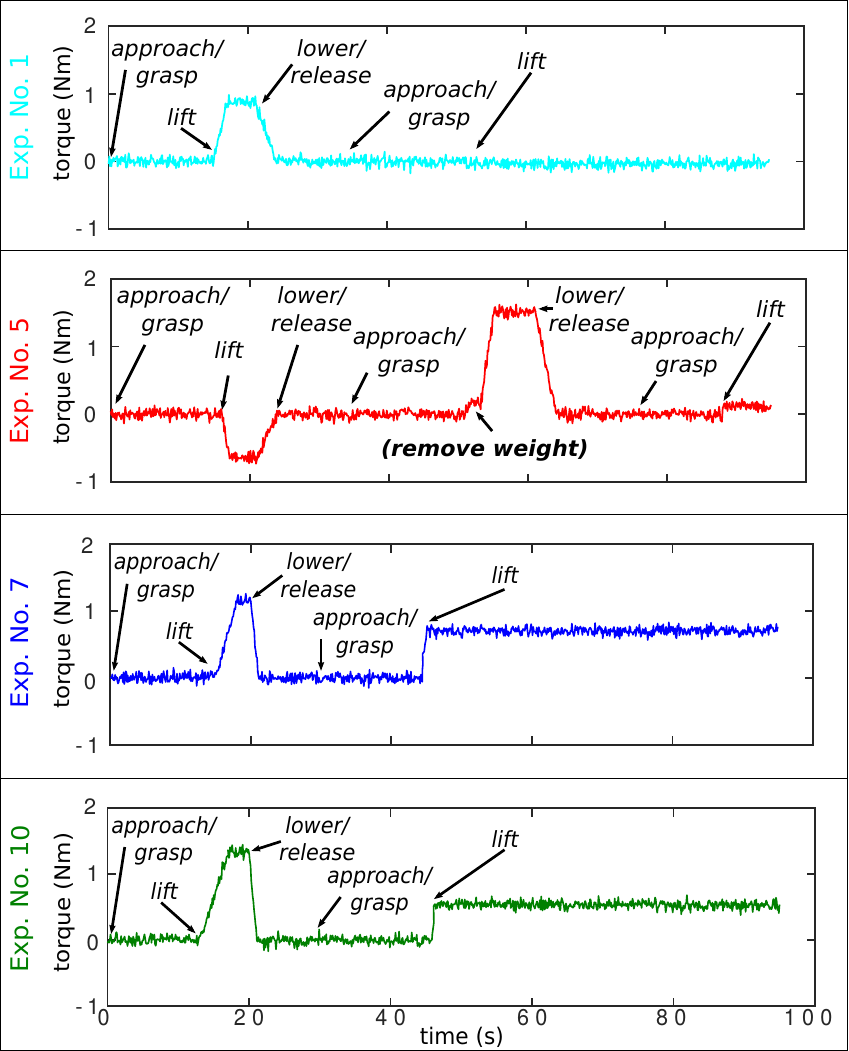}
  \end{center}
  \caption{The continuous wrist torque measurements during the Exps. No. 1, 5, 7, and 10 are presented.  The values correspond to Table~\ref{tab:ExpResult1} and Table~\ref{tab:ExpResult2} ones.  The torque values for the rest of the experiments are omitted since they are similar to the presented ones.  In each sub-graph, the actions of the robot on the object are labelled as:  ``approach/grasp'', ``lift'', and ``lower/release'', to show the torque value changes during each experiment.}
  \label{Fig:expft}
\end{figure*}

With the first five experiments (Exps. Nos. 1--5) we tested the ability of the method to reach the real CoM when this is inside the object.  The particular cylindrical object is everywhere graspable.  From the average results of the first four experiments (Exps. Nos. 1--4) in Table~\ref{tab:ExpResult1} and Figs.~\ref{Fig:expd} and~\ref{Fig:expft}, we first note that the visually computed CoM is always found roughly in the center of the object, which gives a reasonable initial grasp point.  One regrasp using the force/torque perception method was always enough to reach the $d_{thres}$ threshold from the CoM or the $\tau_{thres}$  threshold when it is lifted, making the final torque ($\tau'$) or the displacement ($d'$) very close to zero.  The percentage of torque improvement (last column of Table~\ref{tab:ExpResult1} and Fig.~\ref{Fig:expft}) is mostly high.  Together with the low final torques it means that the second grasp is always more wrist torque efficient and reliable, since the object is grasped very close to the real CoM.  In the fifth experiment (Exp. No. 5) we verified that our method can detect online changes in the distribution of the object mass and automatically regrasp the object at its newly estimated CoM position using the force and torque readings.

In the next four experiments (Exps. Nos. 6--9; Table~\ref{tab:ExpResult2} and Fig.~\ref{Fig:expft}) we tested the ability of the method to reach the closest possible grasp to the real CoM (where the torque is the minimum possible using a single hand), when the CoM is out of the object and the object includes non-graspable areas.  We note that the torque is always minimized after the regrasp, which makes the grasping more torque efficient.  In the tenth experiment we tested the ability of finding the closest grasp to the CoM that is inside the object but in an non-graspable position.  The final grasp is close to the first one.  Note that the slight hand displacement (very close to zero) is due to different set of handle-like grasps localization during the visual stage of the second regrasp iteration.

Last but not least, in the eleventh experiment (Exp. No. 11) we just tested the ability to recognize a negligible torque during the first object lift, since the object (drill) is very small.  This is the reason that a regrasps is not required. In all cases the possible minimum torque was reached after a single regrasp.

\subsection{Discussion and Limitations}
The results of the proposed framework show that the combination of proprioceptive 3D range and exteroceptive force/torque sensing decreases the torque effort at the wrist level, when lifting heavy objects with one hand.  The methodology, as appears in this paper, has some limitation that we discuss briefly in this section.

To start with the proprioceptive 3D range part, the presented method segments an object from the environment, assuming a dominant plane (e.g., a table) that the objects lies on.  The extracted 3D point cloud of the object is then used to estimate visually the object's CoM.  This simplified segmentation method has an obvious limitation in cases where the objects do not lie on a table, e.g., objects in a bin, or when multiple ones are occluded.  Given though that this part acts as a black box in our framework, one can apply any object segmentation method (including object recognition/detection and localization) to extract the required point cloud in real-world environments.  In addition, during voxelization for localizing the CoM, we assumed that the object is made by isotropic material with constant mass density.  Given that for some objects, such as hammers, this assumption may not hold, a more sophisticated technique should be developed to detect different material densities of objects and adapt accordingly the voxelization technique.  Furthermore, the focus of this paper is on objects than include handle-like graspable areas.  As previously mentioned in the related work (Sec.~\ref{Sec:rw}), newly developed methods can localize various types of graspable areas on robots (using for instance deep learning).  We envision a system that includes such methods, but given that a big set of objects have handles, in this paper, we focused primarily on them.

During the exteroceptive force/torque regrasping part, the method minimizes the torque effort at the wrist level of the robot, where the sensor is installed.  In this way, the produced torque in the rest of the robot's joints is not considered.  Moreover, the robot does not change the orientation of the object while lifting it.  These types of maneuvers could minimize the torque effort at the wrist level.  In addition, the method considers only a single arm use, while the framework has more potentials using two arms or even the whole body to minimize the torque effort.  Last but not the least, ``full palm'' grasps limit the introduced framework in the type of hands that can be used.  Other types, such as precision grippers, can introduce slippage during object lifting, resulting in potential torque imprecisions. Thus, the method needs to be extended to monitor torques and slippage during lifting, for instance, with the use of tactile sensing.

Experimentally, the method has been tested on relatively simplified objects, whereas the experiments should be extended to more complex ones (such as chairs), where occlusions and grasping limitations may exist.

\section{Conclusions and Future Works}
In this paper, we presented a novel combination of 3D range and force/torque sensing for finding CoM-based grasps on heavy objects that include handles.  By first applying a visual CoM estimation using point cloud data coming from a range sensor and then applying a set of regrasps to measure the forces and torques on the wrist of the arm, our method is able to accurately detect the real CoM position of the object and grasp it from the most torque efficient handle grasp.  In the experiments, we showed that one regrasp after the visual-based one is enough to localize the most wrist torque efficient one on the humanoid robot WALK-MAN.

In future work, we first plan to improve the visual estimation of the CoM by using a SLAM method like the Moving KinectFusion\cite{RV12} for building a better point cloud representation of the object while the head or the whole robot is moving.  We also plan to generalize our method by using two hands for the manipulation or by considering whole-body motions for more secure bi-manual grasping.  One further extension could be the application of different strategies when torque/force limits are reached during the object lifting phase, since the object may slip or rotate during a hand grasp.

\section*{Acknowledgment}
\addcontentsline{toc}{section}{Acknowledgment}
This work is supported by the FP7-ICT-2013-10 WALK-MAN European Commission project, no. 611832.
\bibliographystyle{amsplain}
\bibliography{ijhr2017}

\providecommand{\bysame}{\leavevmode\hbox to3em{\hrulefill}\thinspace}
\providecommand{\MR}{\relax\ifhmode\unskip\space\fi MR }
\providecommand{\MRhref}[2]{%
  \href{http://www.ams.org/mathscinet-getitem?mr=#1}{#2}
}
\providecommand{\href}[2]{#2}
\begin{thebibliography}{10}

\bibitem{PK11}
Anna A.~Petrovskaya and Oussama Khatib, \emph{{Global Localization of Objects
  via Touch}}, IEEE Transactions on Robotics (T-RO) \textbf{27} (2011), no.~3,
  569--585.

\bibitem{HCDDSD13}
Hussam Al~Hussein, Tiago Caldeira, Dongming Gan, Jorge Dias, and Lakmal
  Seneviratne, \emph{{Object Shape Perception in Blind Robot Grasping using a
  Wrist Force/Torque Sensor}}, IEEE 20th International Conference on
  Electronics, Circuits, and Systems (ICECS) (Abu Dhabi, United Arab Emirates),
  IEEE, 2013, pp.~193--196.

\bibitem{AMOL99}
Peter~K. Allen, Andrew~T. Miller, Paul~Y. Oh, and Brian~S. Leibowitz,
  \emph{{Integration of Vision, Force and Tactile Sensing for Grasping}},
  International Journal of Intelligent Machines \textbf{4} (1999), no.~1,
  129--149.

\bibitem{AAH85}
Christopher~G. Atkeson, Chae~H. An, and John~M. Hollerbach, \emph{{Rigid Body
  Load Identification for Manipulators}}, IEEE 24th Conference on Decision and
  Control (Fort Lauderdale, FL, USA), IEEE, 1985, pp.~996--1002.

\bibitem{BJ52}
John~Perry Ballantine and Arthur~Rudolph Jerbert, \emph{{Distance from a Line
  or Plane to a Point}}, American Mathematical Monthly \textbf{59} (1952),
  no.~4, 242--243.

\bibitem{BYDK11}
Yasemin Bekiroglu, Renaud Detry, and Danica Kragic, \emph{{Learning Tactile
  Characterizations of Object- and Pose-Specific Grasps}}, IEEE/RSJ
  International Conference on Intelligent Robots and Systems (IROS) (San
  Francisco, CA, USA), IEEE/RSJ, 2011, pp.~1554--1560.

\bibitem{BSWK13}
Yasemin Bekiroglu, Dan Song, Lu~Wang, and Danica Kragic, \emph{{A Probabilistic
  Framework for Task-Oriented Grasp Stability Assessment}}, IEEE International
  Conference on Robotics and Automation (ICRA) (Karlsruhe, Germany), IEEE,
  2013, pp.~3040--3047.

\bibitem{BicchiK00}
Antonio Bicchi and Vijay Kumar, \emph{{Robotic Grasping and Contact: a
  Review}}, IEEE International Conference on Robotics and Automation (ICRA)
  (San Francisco, CA, USA), IEEE, 2000, pp.~348--353.

\bibitem{BSAL13}
Joao Bimbo, Lakmal~D. Seneviratne, Kaspar Althoefer, and Hongbin Liu,
  \emph{{Combining Touch and Vision for the Estimation of an Object's Pose
  During Manipulation}}, IEEE/RSJ International Conference on Intelligent
  Robots and Systems (IROS) (Tokyo, Japan), IEEE/RSJ, 2013, pp.~4021--4026.

\bibitem{CGSFPB12}
Manuel~G. Catalano, Giorgio Grioli, Alessandro Serio, Edoardo Farnioli,
  Cristina Piazza, and Antonio Bicchi, \emph{{Adaptive Synergies for a Humanoid
  Robot Hand}}, IEEE-RAS 12th International Conference on Humanoid Robots
  (Humanoids) (Osaka, Japan), IEEE-RAS, 2012, pp.~7--14.

\bibitem{CSF12}
Lillian Chang, Joshua~R. Smith, and Dieter Fox, \emph{{Interactive Singulation
  of Objects from a Pile}}, IEEE International Conference on Robotics and
  Automation (ICRA) (Saint Paul, MN, USA), IEEE, 2012, pp.~3875--3882.

\bibitem{DEMK13}
Renaud Detry, Carl~Henrik Ek, Marianna Madry, and Danica Kragic,
  \emph{{Learning a Dictionary of Prototypical Grasp-Predicting Parts from
  Grasping Experience}}, IEEE International Conference on Robotics and
  Automation (ICRA) (Karlsruhe, Germany), IEEE, 2013, pp.~601--608.

\bibitem{EMW14}
Adrien Escande, Nicolas Mansard, and Pierre-Brice Wieber, \emph{{Hierarchical
  Quadratic Programming: Fast Online Humanoid-Robot Motion Generation}},
  International Journal of Robotics Reasearch (IJRR) \textbf{33} (2014), no.~7,
  1006--1028.

\bibitem{FV12}
David Fischinger and Markus Vincze, \emph{{Empty the Basket - a Shape Based
  Learning Approach for Grasping Piles of Unknown Objects}}, IEEE/RSJ
  International Conference on Intelligent Robots and Systems (IROS) (Vilamoura,
  Portugal), IEEE/RSJ, 2012, pp.~2051--2057.

\bibitem{FVJ13}
David Fischinger, Markus Vincze, and Yun Jiang, \emph{{Learning Grasps for
  Unknown Objects in Cluttered Scenes}}, IEEE International Conference on
  Robotics and Automation (ICRA) (Karlsruhe, Germany), IEEE, 2013,
  pp.~609--616.

\bibitem{Gualtieri16}
Marcus Gualtieri, Andreas ten Pas, Kate Saenko, and Robert Platt, \emph{{High
  Precision Grasp Pose Detection in Dense Clutter}}, IEEE/RSJ International
  Conference on Intelligent Robots and Systems (IROS) (Daejeon, South Korea),
  IEEE/RSJ, 2016, pp.~598--605.

\bibitem{HHMB11}
Paul Hebert, Nicolas Hudson, Jeremy Ma, and Joel Burdick, \emph{{Fusion of
  Stereo Vision, Force-Torque, and Joint Sensors for Estimation of In-Hand
  Object Location}}, IEEE International Conference on Robotics and Automation
  (ICRA) (Shanghai, China), IEEE, 2011, pp.~5935--5941.

\bibitem{HPKRAS12}
Alexander Herzog, Peter Pastor, Mrinal Kalakrishnan, Ludovic Righetti, Tamim
  Asfour, and Stefan Schaal, \emph{{Template-Based Learning of Grasp
  Selection}}, IEEE International Conference on Robotics and Automation (ICRA)
  (Saint Paul, MN, USA), IEEE, 2012, pp.~2379--2384.

\bibitem{HHRB11}
Dirk Holz, Stefan Holzer, Radu~Bogdan Rusu, and Sven Behnke, \emph{{Real-Time
  Plane Segmentation using RGB-D Cameras}}, 15th RoboCup International
  Symposium (Istanbul, Turkey), Lecture Notes in Computer Science, vol. 7416,
  Springer, July 2011, pp.~307--317.

\bibitem{HRDGN12}
Stefan Holzer, Radu~Bogdan Rusu, Michael Dixon, Suat Gedikli, and Nassir Navab,
  \emph{{Adaptive Neighborhood Selection for Real-Time Surface Normal
  Estimation from Organized Point Cloud Data Using Integral Images}}, IEEE/RSJ
  International Conference on Intelligent Robots and Systems (IROS) (Vilamoura,
  Portugal), IEEE/RSJ, 2012, pp.~2684--2689.

\bibitem{HHKM98}
Kyuhei Honda, Tsutomu Hasegawa, Toshihiro Kiriki, and Takeshi Matsuoka,
  \emph{{Real-time Pose Estimation of an Object Manipulated by Multi-Fingered
  Hand using 3D Stereo Vision and Tactile Sensing}}, IEEE/RSJ International
  Conference on Intelligent Robots and Systems (IROS) (Victoria, BC, Canada),
  IEEE/RSJ, 1998, pp.~1814--1819.

\bibitem{JNMS15}
Lorenzo Jamone, Lorenzo Natale, Giorgio Metta, and Giulio Sandini,
  \emph{{Highly Sensitive Soft Tactile Sensors for an Anthropomorphic Robotic
  Hand}}, IEEE Sensors Journal \textbf{15} (2015), no.~8, 4226--4233.

\bibitem{JMS11}
Yun Jiang, Stephen Moseson, and Ashutosh Saxena, \emph{{Efficient Grasping from
  RGBD Images: Learning using a New Rectangle Representation}}, IEEE
  International Conference on Robotics and Automation (ICRA) (Shanghai, China),
  IEEE, 2011, pp.~3304--3311.

\bibitem{Kaiser16}
Peter Kaiser, Eren~E. Aksoy, Markus Grotz, Dimitrios Kanoulas, Nikos~G.
  Tsagarakis, and Tamim Asfour, \emph{{Experimental Evaluation of a Perceptual
  Pipeline for Hierarchical Affordance Extraction}}, 2016 International
  Symposium on Experimental Robotics (ISER), vol.~1, Springer, 2017,
  pp.~136--146.

\bibitem{KKGMRMTA16}
Peter Kaiser, Dimitrios Kanoulas, Markus Grotz, Luca Muratore, Alessio Rocchi,
  Enrico Mingo~Hoffman, Nikos~G. Tsagarakis, and Tamim Asfour, \emph{{An
  Affordance-Based Pilot Interface for High-Level Control of Humanoid Robots in
  Supervised Autonomy}}, IEEE-RAS International Conference on Humanoid Robots
  (Humanoids) (Cancun, Mexico), IEEE-RAS, 2016, pp.~621--628.

\bibitem{Kanoulas14}
Dimitrios Kanoulas, \emph{{Curved Surface Patches for Rough Terrain
  Perception}}, Ph.D. thesis, CCIS, Northeastern University, August 2014.

\bibitem{KLTC17}
Dimitrios Kanoulas, Jinoh Lee, Darwin~G. Caldwell, and Nikos~G. Tsagarakis,
  \emph{{Visual Grasp Affordance Localization in Point Clouds using Curved
  Contact Patches}}, International Journal of Humanoid Robotics (IJHR)
  \textbf{14} (2017), no.~1, 1650028--1--1650028--21.

\bibitem{SPL}
Dimitrios Kanoulas and Marsette Vona, \emph{{The Surface Patch Library (SPL)}},
  IEEE International Conference on Robotics and Automation (ICRA) Workshop:
  MATLAB/Simulink for Robotics Education and Research (Hong Kong), IEEE, 2014,
  \url{dkanou.github.io/projects/spl/}, pp.~1--9.

\bibitem{KKBS13}
Dov Katz, Moslem Kazemi, J.~Andrew Bagnell, and Anthony Stentz, \emph{{Clearing
  a Pile of Unknown Objects using Interactive Perception}}, 2013 IEEE
  International Conference on Robotics and Automation (ICRA) (Karlsruhe,
  Germany), IEEE, 2013, pp.~154--161.

\bibitem{KCY93}
Arie Kaufman, Daniel Cohen, and Roni Yagel, \emph{{Volume Graphics}}, Computer
  \textbf{26} (1993), no.~7, 51--64.

\bibitem{KVBP12}
Moslem Kazemi, Jean-Sebastien Valois, J.~Andrew Bagnell, and Nancy Pollard,
  \emph{{Robust Object Grasping using Force Compliant Motion Primitives}},
  Robotics: Science and Systems (RSS) (Sydney, Australia), MIT Press, 2012,
  pp.~177--185.

\bibitem{KET07}
Charles~C. Kemp, Aaron Edsinger, and Eduardo Torres-Jara, \emph{{Challenges for
  Robot Manipulation in Human Environments [Grand Challenges of Robotics]}},
  IEEE Robotics and Automation Society RAM \textbf{14} (2007), no.~1, 20--29.

\bibitem{KRCGNK11}
Ellen Klingbeil, Deepak Rao, Blake Carpenter, Varun Ganapathi, Andrew~Y. Ng,
  and Oussama Khatib, \emph{{Grasping with Application to an Autonomous
  Checkout Robot}}, IEEE International Conference on Robotics and Automation
  (ICRA) (Shanghai, China), IEEE, 2011, pp.~2837--2844.

\bibitem{KP14}
Oliver Kroemer and Jan Peters, \emph{{Predicting Object Interactions from
  Contact Distributions}}, IEEE/RSJ International Conference on Intelligent
  Robots and Systems (IROS) (Chicago, IL, USA), IEEE/RSJ, 2014, pp.~3361--3367.

\bibitem{LNK12}
Joanna Laaksonen, Ekaterina Nikandrova, and Ville Kyrki, \emph{{Probabilistic
  Sensor-Based Grasping}}, IEEE/RSJ International Conference on Intelligent
  Robots and Systems (IROS) (Vilamoura, Portugal), IEEE/RSJ, 2012,
  pp.~2019--2026.

\bibitem{LaValle06}
Steven~M. LaValle, \emph{Planning algorithms}, Cambridge University Press, New
  York, NY, USA, 2006.

\bibitem{LLS15}
Ian Lenz, Honglak Lee, and Ashutosh Saxena, \emph{{Deep Learning for Detecting
  Robotic Grasps}}, The International Journal of Robotics Research (IJRR),
  Special Issue on Robot Vision \textbf{34} (2015), no.~4-5, 705--724.

\bibitem{LPKQ16}
Sergey Levine, Peter Pastor, Alex Krizhevsky, and Deirdre Quillen,
  \emph{{Learning Hand-Eye Coordination for Robotic Grasping with Deep Learning
  and Large-Scale Data Collection}}, CoRR \textbf{abs/1603.02199} (2016).

\bibitem{LBKB14}
Miao Li, Yasemin Bekiroglu, Danica Kragic, and Aude Billard, \emph{{Learning of
  Grasp Adaptation through Experience and Tactile Sensing}}, IEEE/RSJ
  International Conference on Intelligent Robots and Systems (IROS) (Chicago,
  IL, USA), IEEE/RSJ, 2014, pp.~3339--3346.

\bibitem{LPYtPRSA14}
Rui Li, Robert Platt~Jr., Wenzhen Yuan, Andreas ten Pas, Nathan Roscup,
  Mandayam~A. Srinivasan, and Edward Adelson, \emph{{Localization and
  Manipulation of Small Parts using GelSight Tactile Sensing}}, IEEE/RSJ
  International Conference on Intelligent Robots and Systems (IROS) (Chicago,
  IL, USA), IEEE/RSJ, 2014, pp.~3988--3993.

\bibitem{LDSA05}
Efrain Lopez-Damian, Daniel Sidobre, and Rachid Alami, \emph{{A Grasp Planner
  Based On Inertial Properties}}, IEEE International Conference on Robotics and
  Automation (ICRA) (Barcelona, Spain), IEEE, 2005, pp.~754--759.

\bibitem{Mar2015}
Tanis Mar, Vadim Tikhanoff, Giorgio Metta, and Lorenzo Natale,
  \emph{{Self-supervised Learning of Grasp Dependent Tool Affordances on the
  iCub Humanoid Robot}}, IEEE International Conference on Robotics and
  Automation (ICRA) (Seattle, WA, USA), IEEE, 2015, pp.~3200--3206.

\bibitem{MS85}
Matthew~T. Mason and J.~Kenneth Salisbury~Jr., \emph{{Robot Hands and the
  Mechanics of Manipulation}}, MIT Press, 1985.

\bibitem{MFN06}
Giorgio Metta, Paul Fitzpatrick, and Lorenzo Natale, \emph{{YARP: Yet Another
  Robot Platform}}, International Journal on Advanced Robotics Systems
  \textbf{3} (2006), no.~1, 43--48.

\bibitem{RHCT15}
Enrico Mingo~Hoffman, Alessio Rocchi, Arturo Laurenzi, and Nikos~G. Tsagarakis,
  \emph{{Robot control for dummies: Insights and examples using OpenSoT}},
  IEEE-RAS International Conference on Humanoid Robots (Humanoids) (Birmingham,
  UK), IEEE-RAS, 2017, pp.~736--741.

\bibitem{MLBS08}
Luis Montesano, Manuel Lopes, Alexandre Bernardino, and Jose Santos-Victor,
  \emph{{Learning Object Affordances: From Sensory--Motor Coordination to
  Imitation}}, IEEE Transactions on Robotics (T-RO) \textbf{24} (2008), no.~1,
  15--26.

\bibitem{MLHRCT17}
Luca Muratore, Arturo Laurenzi, Enrico Mingo~Hoffman, Alessio Rocchi, Darwin~G.
  Caldwell, and Nikos~G. Tsagarakis, \emph{{XBotCore: A Real-Time Cross-Robot
  Software Platform}}, IEEE International Conference on Robotic Computing (IRC)
  (Taichung, Taiwan), IEEE, 2017, pp.~77--80.

\bibitem{MSZ94}
Richard~M. Murray, S.~Shankar Sastry, and Li~Zexiang, \emph{{A Mathematical
  Introduction to Robotic Manipulation}}, CRC Press, Inc.Boca Raton, FL, USA,
  1994.

\bibitem{MTFA15}
Austin Myers, Ching~L. Teo, Cornelia Fermuller, and Yiannis Aloimonos,
  \emph{{Affordance Detection of Tool Parts from Geometric Features}}, IEEE
  International Conference on Robotics and Automation (ICRA) (Seattle, WA,
  USA), IEEE, 2015, pp.~3200--3206.

\bibitem{Nguyen16}
Anh Nguyen, Dimitrios Kanoulas, Darwin~G. Caldwell, and Nikos~G. Tsagarakis,
  \emph{{Detecting Object Affordances with Convolutional Neural Networks}},
  IEEE/RSJ International Conference on Intelligent Robots and Systems (IROS)
  (Daejeon, South Korea), IEEE/RSJ, 2016, pp.~2765--2770.

\bibitem{Nguyen16b}
\bysame, \emph{{Preparatory Object Reorientation for Task-Oriented Grasping}},
  IEEE/RSJ International Conference on Intelligent Robots and Systems (IROS)
  (Daejeon, South Korea), IEEE/RSJ, 2016, pp.~893--899.

\bibitem{Nguyen17}
\bysame, \emph{{Object-Based Affordances Detection with Convolutional Neural
  Networks and Dense Conditional Random Fields}}, IEEE/RSJ International
  Conference on Intelligent Robots and Systems (IROS) (Vancouver, BC, Canada),
  IEEE/RSJ, 2017, pp.~5908--5915.

\bibitem{OC01}
Allison~M. Okamura and Mark~R. Cutkosky, \emph{{Feature Detection for Haptic
  Exploration with Robotic Fingers}}, The International Journal of Robotics
  Research (IJRR) \textbf{20} (2001), no.~12, 925--938.

\bibitem{Petrovskaya2016}
Anna Petrovskaya and Kaijen Hsiao, \emph{{Active Manipulation for Perception}},
  pp.~1037--1062, Springer International Publishing, Cham, 2016.

\bibitem{PKTN06}
Anna Petrovskaya, Oussama Khatib, Sebastian Thrun, and Andrew~Y. Ng,
  \emph{{Bayesian Estimation for Autonomous Object Manipulation Based on
  Tactile Sensors}}, IEEE International Conference on Robotics and Automation
  (ICRA) (Orlando, FL, USA), IEEE, 2006, pp.~707--714.

\bibitem{PK13}
Alessandro Pieropan, Carl~Henrik Ek, and Hedvig Kjellstr{\"o}m,
  \emph{{Functional Object Descriptors for Human Activity Modeling}}, IEEE
  International Conference on Robotics and Automation (ICRA) (Karlsruhe,
  Germany), IEEE, 2013, pp.~1282--1289.

\bibitem{PG16}
Lerrel Pinto and Abhinav Gupta, \emph{{Supersizing Self-supervision: Learning
  to Grasp from 50K Tries and 700 Robot Hours}}, IEEE International Conference
  on Robotics and Automation (ICRA) (Stockholm, Sweden), IEEE, 2016,
  pp.~3406--3413.

\bibitem{RV12}
Henry Roth and Marsette Vona, \emph{{Moving Volume KinectFusion}}, British
  Machine Vision Conference (BMVC) (Surrey, UK), BMVA Press, Sept. 2012,
  pp.~1--11.

\bibitem{RC11}
Radu~Bogdan Rusu and Steve Cousins, \emph{{3D is here: Point Cloud Library
  (PCL)}}, IEEE International Conference on Robotics and Automation (ICRA)
  (Shanghai, China), IEEE, 2011, pp.~1--4.

\bibitem{SPVFHRMTB14}
Alessandro Settimi, Corrado Pavan, Valerio Varricchio, Mirko Ferrati, Enrico
  Mingo~Hoffman, Alessio Rocchi, Kamilo Melo, Nikos~G. Tsagarakis, and Antonio
  Bicchi, \emph{{A Modular Approach for Remote Operation of Humanoid Robots in
  Search and Rescue Scenarios}}, International Workshop on Modelling and
  Simulation for Autonomous Systems (MESAS), vol. 8906, Springer, Switzerland,
  2014, pp.~192--205.

\bibitem{PMPKRB08}
Michael Stark, Philipp Lies, Michael Zillich, Jeremy Wyatt, and Bernt Schiele,
  \emph{{Functional Object Class Detection Based on Learned Affordance Cues}},
  6th International Conference Computer Vision Systems (ICVS) (Berlin,
  Heidelberg), Springer, 2008, pp.~435--444.

\bibitem{SNK07}
Mike Stilman, Koichi Nishiwaki, and Satoshi Kagami, \emph{{Learning Object
  Models for Whole Body Manipulation}}, IEEE-RAS 7th International Conference
  on Humanoid Robots (Humanoids) (Pittsburgh, PA, USA), IEEE-RAS, 2007,
  pp.~174--179.

\bibitem{SSHB13}
J\"org St\"uckler, Ricarda Steffens, Dirk Holz, and Sven Behnke,
  \emph{{Efficient 3D Object Perception and Grasp Planning for Mobile
  Manipulation in Domestic Environments}}, Robotics and Autonomous Systems
  (RAS) \textbf{61} (2013), no.~10, 1106--1115.

\bibitem{tPP14}
Andreas ten Pas and Robert Platt, \emph{{Localizing Grasp Affordances in 3-D
  Points Clouds Using Taubin Quadric Fitting}}, International Symposium on
  Experimental Robotics (ISER) (Marrakech and Essaouira, Morocco), Springer,
  2014.

\bibitem{tPP15}
\bysame, \emph{{Using Geometry to Detect Grasps in 3D Point Clouds}}, The
  International Symposium on Robotics Research (ISRR) (Sestri Levante, Italy),
  Springer, 2015.

\bibitem{Tsagarakis2016}
Nikos~G. Tsagarakis, D.~G. Caldwell, F.~Negrello, W.~Choi, L.~Baccelliere, V.G.
  Loc, J.~Noorden, L.~Muratore, A.~Margan, A.~Cardellino, L.~Natale,
  E.~Mingo~Hoffman, H.~Dallali, N.~Kashiri, J.~Malzahn, J.~Lee, P.~Kryczka,
  D.~Kanoulas, M.~Garabini, M.~Catalano, M.~Ferrati, V.~Varricchio,
  L.~Pallottino, C.~Pavan, A.~Bicchi, A.~Settimi, A.~Rocchi, and A.~Ajoudani,
  \emph{{WALK-MAN: A High-Performance Humanoid Platform for Realistic
  Environments}}, Journal of Field Robotics (JFR) \textbf{34} (2017), no.~7,
  1225--1259.

\bibitem{VV12}
Karthik~Mahesh Varadarajan and Markus Vincze, \emph{{AfRob: The Affordance
  Network Ontology for Robots}}, IEEE/RSJ International Conference on
  Intelligent Robots and Systems (IROS) (Vilamoura, Portugal), IEEE/RSJ, 2012,
  pp.~1343--1350.

\bibitem{VPBCN17}
Giulia Vezzani, Ugo Pattacini, Giorgio Battistelli, Luigi Chisci, and Lorenzo
  Natale, \emph{{Memory Unscented Particle Filter for 6-DOF Tactile
  Localization}}, IEEE Transactions on Robotics (T-RO) \textbf{33} (2017),
  no.~5, 1139--1155.

\bibitem{VBSKK13}
Francisco Vina, Yasemin Bekiroglu, Christian Smith, Yiannis Karayiannidis, and
  Danica Kragic, \emph{{Predicting Slippage and Learning Manipulation
  Affordances through Gaussian Process Regression}}, IEEE-RAS International
  Conference on Humanoid Robots (Humanoids) (Atlanta, GA, USA), IEEE-RAS, 2013,
  pp.~462--468.

\bibitem{YBA11}
Hanna Yousef, Mehdi Boukallel, and Kaspar Althoefer, \emph{{Tactile Sensing for
  Dexterous In-Hand Manipulation in Robotics --- A Review}}, Sensors and
  Actuators A: Physical \textbf{167} (2011), no.~2, 171 -- 187.

\bibitem{ZLT13}
Li~Zhang, Siwei Lyu, and Jeff Trinkle, \emph{{A Dynamic Bayesian Approach to
  Real Time Estimation and Filtering in Grasp Acquisition}}, IEEE International
  Conference on Robotics and Automation (ICRA) (Karlsruhe, Germany), IEEE,
  2013, pp.~85--92.

\end{thebibliography}

\noindent%
\parbox{5truein}{
\begin{minipage}[b]{1truein}
\centerline{{\psfig{file=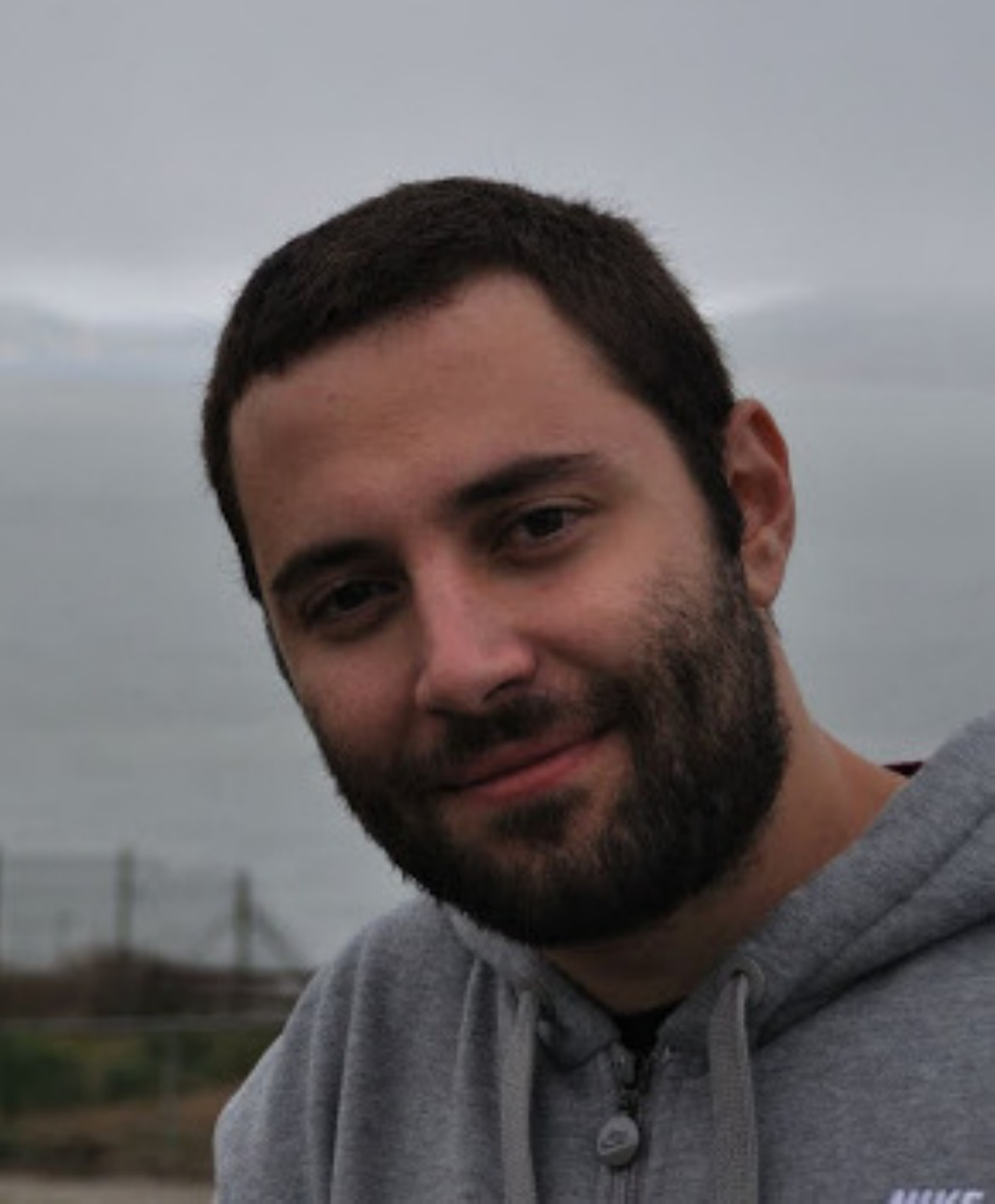,width=1in,height=1.25in}}}
\end{minipage}
\hfill         
\begin{minipage}[b]{3.85truein}
{{\bf Dimitrios Kanoulas} is a Postdoctoral Researcher at the Humanoids $\&$ Human Centered Mechatronics (HHCM) lab at the Istituto Italiano di Tecnologia (IIT), working with Dr. Nikos Tsagarakis and Prof. Darwin Caldwell on the perception and learning part of the COMAN, WALK-MAN, CENTAURO, and CogIMon EU-funded research projects.  He was the perception team leader during the DARPA Robotics Challenge Finals in
\hfilneg}
\end{minipage} } 

\vspace*{4.8pt}
\noindent
2015.  In August 2014, he completed his Ph.D. at the Geometric and Physical Computing (GPC) Lab in the College of Computer and Information Science at Northeastern University, advised by Prof. Marsette Vona. He started out as a member of the Algorithms and Theory group at Northeastern University, advised by Prof. Rajmohan Rajaraman and Ravi Sundaram. During the summer of 2012, he completed an internship at INRIA in France, advised by Prof. Christian Laugier and Dr. Alexandros Makris. He received his B.S. in Computer Engineering and Informatics Department from University of Patras, Greece in 2008, advised by Prof. Paul Spirakis and Dr. Haralampos Tsaknakis. 

\vspace*{13pt}
\noindent%
\parbox{5truein}{
\begin{minipage}[b]{1truein}
\centerline{{\psfig{file=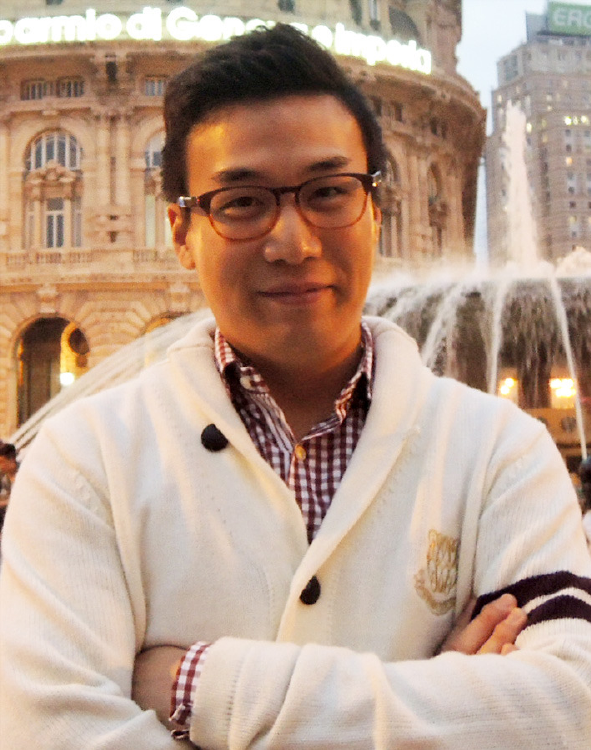,width=1in,height=1.25in}}}
\end{minipage}
\hfill         
\begin{minipage}[b]{3.85truein}
{{\bf Jinoh Lee} received the B.S. degree in Mechanical Engineering from Hanyang University, Seoul, South Korea, in 2003 and the M.Sc. and the Ph.D. degrees in Mechanical Engineering from Korea Advanced Institute of Science and Technology (KAIST), Daejeon, South Korea, in 2012.  Since 2012, he has joined the Department of Advanced Robotics (ADVR), Istituto Italiano di Tecnologia (IIT), Genoa, Italy, as a Postdoctoral Researcher,
\hfilneg}
\end{minipage} } 

\vspace*{4.8pt}
\noindent
involved in projects such as Safe and Autonomous Physical Human-Aware Robot Interaction (SAPHARI) and Whole-body Adaptive Locomotion and Manipulation (WALK-MAN) funded under the European Community's 7th Framework Programme.  He is currently a Research Scientist in ADVR, IIT.
His professional is about robotics and control engineering which include whole-body manipulation of high degrees-of-freedom humanoids, robust control of highly nonlinear systems, and compliant robotic system control for safe human-robot interaction. Since 2014, Dr. Lee has participated in Technical Committee member of International Federation of Automatic Control (IFAC), TC4.3 Robotics.

\vspace*{13pt}  
\noindent%
\parbox{5truein}{
\begin{minipage}[b]{1truein}
\centerline{{\psfig{file=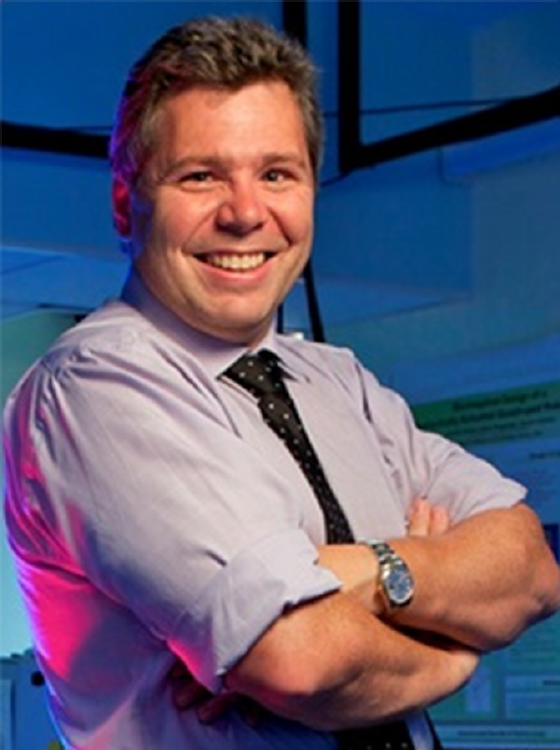,width=1in,height=1.25in}}}
\end{minipage}
\hfill 
\begin{minipage}[b]{3.85truein}

{{\bf Darwin G. Caldwell} received his B.Sc. and Ph.D. degrees in Robotics from the University of Hull, in 1986 and 1990, respectively and an M.Sc. in Management in 1996 from the University of Salford.  He is currently a Research Director at the Italian Institute of Technology, Genoa, Italy, and an Honorary Professor at the Universities of Sheffield, Manchester, Kings College London, Bangor, and Tianjin (China).  He is the author or co-author of 
\hfilneg}
\end{minipage} } 

\vspace*{4.8pt}
\noindent
more than 500 academic papers, and has 15 patents.  His research interests include innovative actuators, force augmentation exoskeletons, humanoid (iCub, cCub, COMAN and WalkMan) and quadrupled robots (HyQ and HyQ2Max), and medical robotics.  In 2015, he was elected a Fellow of the Royal Academy of Engineering.

\vspace*{13pt}  
\noindent%
\parbox{5truein}{
\begin{minipage}[b]{1truein}
\centerline{{\psfig{file=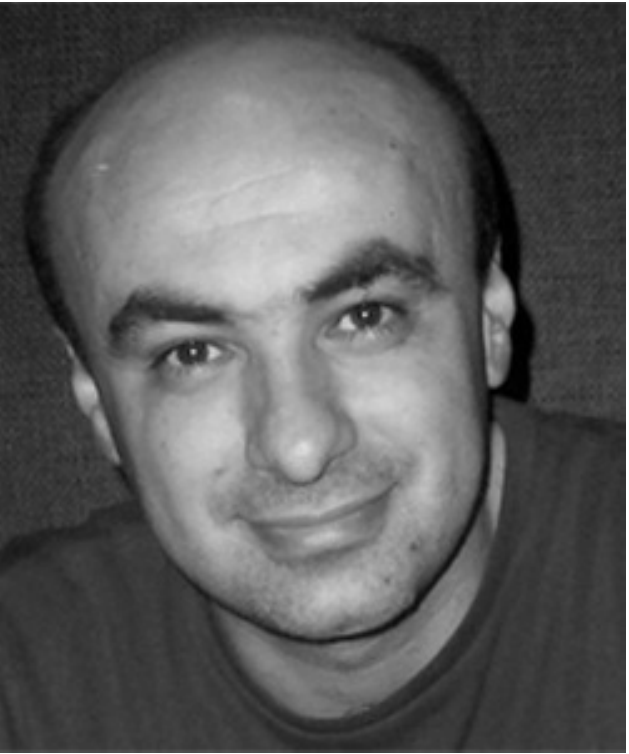,width=1in,height=1.25in}}}
\end{minipage}
\hfill 
\begin{minipage}[b]{3.85truein}
{{\bf Nikos G. Tsagarakis} received his D.Eng. degree in Electrical and Computer Science Engineering in 1995 from the Polytechnic School of Aristotle University, Greece, an M.Sc. degree in Control Engineering in 1997 and in 2000 a Ph.D. in Robotics from the University of Salford, UK.  He is a Tenured Senior Scientist at IIT with overall responsibility for Humanoid design $\&$ Human Centred Mechatronics development.  He is an author or co-author
\hfilneg}
\end{minipage} } 

\vspace*{4.8pt}
\noindent
 of over 250 papers in journals and at international conferences and holds 14 patents.  He has received the Best Jubilee Video Award at IROS (2012), the 2009 PE Publishing Award from the Journal of Systems and Control Engineering and prizes for Best Paper at ICAR (2003) and the Best Student Paper Award at Robio (2013). He was also a Finalist for Best Entertainment Robots and Systems-20th Anniversary Award at IROS (2007) and finalist for the Best Manipulation paper at ICRA (2012), the Best Conference Paper at Humanoids (2012), the Best Student Papers at Robio (2013) and ICINCO (2014). He has been in the Program Committee of over 60 international conferences including ICRA, IROS, RSS, HUMANOIDS, BIOROB and ICAR, and he is a Technical Editor of IEEE/ASME Transactions in Mechatronics and on the Editorial Board of Robotics and Automation Letters. Since 2013, he is also serving as a Visiting Professor at the Centre for Robotics Research (CORE), Department of Informatics, King’s College University, London, UK.

\vfill\eject

\end{document}